\documentclass[final]{clv2025}

\jvol{vv}
\jnum{nn}
\jyear{2026}
\dochead{Article}
\pageonefooter{}

\usepackage{amsmath}
\usepackage{amssymb}
\usepackage{booktabs}
\usepackage{longtable}
\usepackage{array}
\usepackage{calc}
\usepackage{graphicx}
\usepackage{url}
\usepackage{microtype}
\hypersetup{hidelinks,
  pdfauthor={Rafael da Silva and Jeff Eicher},
  pdftitle={Measuring the Cross-Lingual Comprehension Gap: How the language of the evidence shapes what language models understand}}
\newcommand{\pandocbounded}[1]{#1}

\runningtitle{Measuring the Cross-Lingual Comprehension Gap}
\runningauthor{da Silva and Eicher}

\begin{document}

\title{Measuring the Cross-Lingual Comprehension Gap: How the language of the evidence shapes what language models understand}
\author{Rafael da Silva\thanks{Corresponding author}$^{,1}$, Jeff Eicher$^{1}$}
\affilblock{\affil{PhD in Applied Data Science, Eastern University\\
Corresponding author: \email{rafael.dasilva@eastern.edu}\\
Contributing author: \email{jeff.eicher@eastern.edu}}}
\maketitle

\begin{abstract}
Language models are often evaluated as though the capabilities they demonstrate in English remain equally available when the same content is presented in other languages. However, traditional multilingual benchmarks rarely isolate the effect of language while holding constant the content, question, reference answer, model, and unit of evaluation. To measure this loss under controlled conditions, we define the \textbf{Cross-Lingual Comprehension Gap (CLCG)} as the reduction in response quality when the same content and question are presented in a target language rather than in the English reference condition.

The full aligned panel comprised \textbf{559 unique English source articles} and \textbf{9,843 materialized article--language versions} across a panel of 18 languages, corresponding to \textbf{410,186 aligned paragraph--language instances}, \textbf{23,276 unique English body paragraphs}, and \textbf{23,091 unique questions}. The primary confirmatory execution used a stratified sample of \textbf{150 articles}. The language panel included English as the reference language, Portuguese as an empirical high-resource non-English baseline, and 16 target languages spanning Joshi et al.'s (2020) resource classes 0--4; no analyzed target language belonged to Class 5, and Ayacucho Quechua remained unclassified in the primary mapping. Five models from distinct laboratories answered the same questions under the available language conditions. The parallel within-item design held constant the content, question, reference answer, model, prompt, and unit of evaluation, varying only the language of the passage. The executed primary estimator contrasted the English scored-row micro-mean with the pooled target-language scored-row micro-mean using Token-F1 on higher-complexity open-ended questions; confidence intervals were obtained through bootstrap resampling clustered at the article level.

The primary pooled scored-row CLCG relative to English was \textbf{0.078} (95\% CI: \textbf{0.072--0.084}), corresponding to an approximately \textbf{17\% reduction relative to the English reference score}. The equal-language macro summary across the 16 target languages was \textbf{0.077}. Using Portuguese as an empirical high-resource non-English baseline, the equal-language macro net gap was \textbf{0.016} (95\% CI: \textbf{0.013--0.020}). Under the canonical Joshi et al.~(2020) classification, language-level CLCG was negatively associated with resource class (\textbf{\ensuremath{\rho} = \ensuremath{-}0.594}, \textbf{p = 0.015}, \textbf{n = 16}). Thus, lower-resource classes showed larger gaps. Assigning Ayacucho Quechua to Class 1 in a sensitivity analysis produced a similar result (\textbf{\ensuremath{\rho} = \ensuremath{-}0.632}, \textbf{p = 0.007}, \textbf{n = 17}).

In blinded paired human evaluations, responses produced under higher-resource language conditions were preferred in \textbf{61.6\%} of decisive judgments. After accounting for repeated evaluations by annotator and response pair, the estimated probability of preference was \textbf{0.655} (95\% CI: \textbf{0.558--0.741}). Although absolute adequacy ratings showed low item-level inter-rater agreement, the paired evaluation independently supported the relative direction of the effect.

Together, these findings show that capabilities demonstrated in English should not be assumed to remain equally available in other languages. English-centered evaluations may therefore overestimate the quality delivered to users of low-resource languages.
\end{abstract}

\subsection{Introduction}\label{introduction}

A language model's capabilities are not necessarily preserved when only the language of the evidence changes. A system that answers a question correctly after reading a passage in English may produce an incomplete or incorrect response when given the same content in another language, even when the question, expected answer, and remainder of the protocol are unchanged. This asymmetry is especially consequential for languages with fewer digital resources. Even when models are described as multilingual, fully aligned benchmarks show substantially lower and more uneven comprehension or question-answering performance when equivalent inputs are presented in many medium- and low-resource languages rather than in English (Bandarkar et al., 2024; Longpre et al., 2021).

The multilingual evaluation literature extensively documents systematic performance differences across languages (Bandarkar et al., 2024; Zhang et al., 2023). However, cross-language benchmark comparisons can conflate language effects with differences in item difficulty, domain composition, translation, response format, required output language, and possible prior exposure to benchmark content (Longpre et al., 2021; Sánchez Salido et al., 2026). Consequently, a difference in score does not necessarily indicate how much of the observed loss is specifically associated with the language of the text that the model must comprehend. A contrast is therefore needed in which the item, task, model, reference answer, and evaluation protocol are held constant while only the language of the evidence varies.

We formalize this contrast as the \emph{Cross-Lingual Comprehension Gap} (CLCG). The CLCG represents the loss, or preservation, of a model's response quality when the same evidence is presented in a target language rather than in an English reference condition. We therefore report the cross-lingual gap alongside, rather than in place of, per-condition metrics such as accuracy, exact match, Token-F1, and semantic similarity (Adlakha et al., 2024; Bulian et al., 2022). These metrics score each response within a condition, whereas the executed primary CLCG estimator contrasts the English scored-row micro-mean with the pooled target-language scored-row micro-mean to estimate relative preservation of contextual comprehension. The benchmark used in this study therefore functions as an instrument for observing the CLCG rather than as its definition.

To estimate this contrast, we constructed a \emph{within-item} design in which the same model answers the same question using the same English reference answer, defined as the answer treated as correct for evaluation, after reading aligned versions of the same content. Model responses are also always generated in English so that target-language generation ability is not conflated with passage comprehension. The study covers English, Portuguese as a high-resource baseline, and 16 target languages, evaluated using five models from five laboratories. We use a stratified sample of 150 articles from the 559-article aligned panel of ParallelQA-18, a professionally human-translated parallel corpus aligned at the paragraph level (da Silva \& Eicher, 2026b). Drawing on established reading-comprehension taxonomies, we organize questions by progressively broader processing demands: direct recognition of explicitly stated information, reorganization or integration across the text, inference beyond what is explicitly stated, and higher-order interpretation or evaluation, including questions about authorial purpose (\texttt{L0}--\texttt{L5}; Day \& Park, 2005; Yüceer, 2022). The evaluation combines two item sources: questions generated through a controlled procedure and pre-existing human-authored questions originally created to guide readers' comprehension of the texts.

The results reveal a systematic gap. In the primary \texttt{L3}--\texttt{L5} band, the mean score was \texttt{0.456} in English and \texttt{0.395} in Portuguese, yielding an English--Portuguese gap of \texttt{0.061}. The primary pooled scored-row CLCG was \texttt{0.078} (95\% CI \texttt{{[}0.072,\ 0.084{]}}) and remained positive for all five models; the equal-language macro summary across the 16 target languages was \texttt{0.077}. Under the canonical Joshi et al.~(2020) classification, the primary association between resource class and language-level CLCG was negative (\texttt{\ensuremath{\rho}=-0.594}, \texttt{p=0.015}, \texttt{n=16}), as expected because higher class numbers denote greater resource availability. Class-level means were descriptive and unevenly supported: Class 0 contained only Fon, Class 5 contained no analyzed target language, and Ayacucho Quechua remained unclassified in the primary mapping. Swahili, for example, exhibited a smaller gap than the Portuguese baseline.

The evidence also indicates that the phenomenon cannot be attributed to a single weakness in the instrument. The CLCG remained positive for articles published after each model's declared or estimated knowledge cutoff. Declared knowledge cutoffs or documented pretraining time spans can be used as rough proxies for the latest period likely represented in a model's public training data, but they do not establish whether any specific document was included or excluded (Desai et al., 2024; Dhingra et al., 2022). The gap was slightly larger for these later articles than for content published before the corresponding cutoffs. This classification cannot prove the presence or absence of an article in the training data, but it reduces the plausibility that the findings are explained solely by memorization. The effect was also stable across repeated inference runs, showed little sensitivity to the rule used for response-processing failures, and was not explained by differences in structural coverage among translations. Moreover, the ranking of languages was reproduced in an external evaluation based on FLORES-200 (\texttt{\ensuremath{\rho}\ensuremath{\approx}0.79}), and pre-existing human-authored questions yielded a larger CLCG (\texttt{0.152}) than generated items in the primary band (\texttt{0.078}), showing that the direction of the effect did not depend exclusively on automatic question generation.

A separate human-validation study provided partial convergence. Automatic metrics were positively but modestly correlated with human adequacy (\texttt{\ensuremath{\rho}=0.232}) and fact recovery (\texttt{\ensuremath{\rho}=0.285}), each across \texttt{n=564} items with matched post-QC scores; responses derived from higher-resource contexts were preferred in 61.6\% of decisive paired judgments; and class-level human adequacy was directionally compatible with the automatic gap but imprecise (\texttt{\ensuremath{\rho}=-0.600}, \texttt{p=0.285}, \texttt{n=5}). Inter-rater agreement was nevertheless low across all dimensions (\texttt{\ensuremath{\alpha}\ensuremath{\leq}0.154}), limiting the confirmatory strength of this evidence.

The results do not, however, support a simple account in which the gap increases continuously with cognitive depth or can be explained by a single linguistic property. The profile across the six comprehension levels was nonmonotonic: the largest contrasts appeared in literal-extraction and reorganization tasks rather than in the deepest integrative tasks. Among the factors examined, resource availability was the most consistent correlate. Tokenization fertility, use of a non-Latin writing system, and typological isolate status showed no conclusive associations. The CLCG should therefore be interpreted as a systematic and structured inequality that remains heterogeneous across languages, models, and item types.

This study makes four main contributions. First, it formalizes the CLCG as a construct of relative preservation in contextual comprehension and defines an aggregate scored-row estimator embedded in a controlled parallel design, distinguishing the construct from absolute performance scores. Second, it develops a controlled observational design spanning 18 languages, five models, and two question sources, using a professionally human-translated parallel corpus that spans multiple Joshi et al.~(2020) resource classes. Third, it provides an empirical characterization of the magnitude and heterogeneity of the gap, including its relationships with resource availability, models, and comprehension demands. This characterization is supported by a chain of validity evidence combining human-authored items, partial human triangulation of the automatic metrics, likely unseen content, stability across repeated runs, controls for response-processing failures, a context-scope ablation, structural-coverage checks, and cross-domain reproduction using FLORES-200. Fourth, it releases reproducible and extensible research infrastructure through ParallelQA-18 (da Silva \& Eicher, 2026b) and the ParallelQA-18 Builder (da Silva \& Eicher, 2026a). The dataset record contains the releasable research bases---including reconstruction manifests and identifiers, versioned prompts, the generated-question bank, frozen model responses, item-level scores, and raw and processed deidentified human-validation data---but does not redistribute copyrighted article bodies or source-authored question text. The separate Builder reconstructs those protected source materials locally, and the paper-facing analysis code consumes both surfaces to regenerate the analytical source of truth. Together, this infrastructure enables researchers to reproduce the study, apply new metrics to the existing responses, and compare future models under the same protocol without repeating the complete set of original model calls.

The remainder of the paper reviews prior work on multilingual evaluation and comprehension measurement, formally defines the CLCG and its estimator, describes the experimental design, presents the results and their validity and robustness analyses, and discusses the construct's implications and limitations.

\subsection{Related Work}\label{related-work}

\subsubsection{Multilingual Evaluation, Comparability, and Cross-Lingual Validity}\label{multilingual-evaluation-comparability-and-cross-lingual-validity}

Evaluations of language models across multiple languages have repeatedly shown that capabilities demonstrated in English are not uniformly preserved in other languages (Jin et al., 2024; Zhang et al., 2023). Multilingual benchmarks often report higher performance in high-resource languages and larger losses in languages with less digital representation and less pretraining or task-specific data (Goyal et al., 2022; Li et al., 2025). Cross-lingual performance disparities have been observed in both proprietary and open-weight language models, including models evaluated or deployed for multilingual use (Kwak \& Pardos, 2024).

Broader linguistic coverage has made this problem more visible. Benchmarks such as FLORES-200 and Belebele substantially expanded multilingual evaluation beyond high-resource European languages by providing broad, quality-controlled coverage of low-resource and non-European languages (Bandarkar et al., 2024; NLLB Team, 2024). Aggregate multilingual scores can nevertheless obscure substantial performance losses in individual languages, particularly those with low or very low levels of digital and training-data representation (Bandarkar et al., 2024; Goyal et al., 2022). This literature establishes the existence of multilingual inequality, although most findings are reported as performance differences within particular tasks.

Comparing performance across languages, however, requires more than translating a benchmark. Translation quality, item naturalness, task difficulty, domain and content selection, cultural and topical suitability, and document scope can affect cross-language benchmark scores independently of model language ability (Clark et al., 2020; Goyal et al., 2022). Translation can also alter item difficulty through changes in syntactic complexity, source-language priming, translationese, and increased lexical overlap, even in otherwise parallel multilingual datasets (Clark et al., 2020; Liu \& Afzaal, 2021). Comparisons based on different item sets additionally conflate language effects with differences in content and difficulty. Paired multilingual designs using aligned versions of the same item reduce confounding from item content and difficulty when estimating input-language effects (Bandarkar et al., 2024; Huang et al., 2025).

The language of the response introduces another source of confounding. When models must both process information and produce the answer in the target language, the final score conflates task-solving success with the model's ability to realize or translate the answer into that language (Bafna et al., 2025). A poor response may therefore reflect a comprehension failure, a production difficulty, unfavorable tokenization, or an unsuitable evaluation metric. Translate-test, translate-train, and prompt-language choices can materially change multilingual evaluation results (Conneau et al., 2018; Horbach et al., 2024; Jin et al., 2024).

The present study begins from this established phenomenon but asks a more controlled question: how much of the capability demonstrated by a model in English is preserved when only the language of the evidence-bearing text changes? The CLCG design holds constant the model, question, reference answer, semantic content, and evaluation protocol while varying only the language of the passage. This formulation shifts the focus from a general comparison of scores to a controlled cross-language contrast of relative preservation.

\subsubsection{Reading Comprehension, Item Depth, and Evaluation Metrics}\label{reading-comprehension-item-depth-and-evaluation-metrics}

Machine reading comprehension benchmarks differ not only in domain but also in the type of comprehension required. Question-answering items vary substantially in cognitive demand, from local lexical matching and extraction to multi-sentence inference and cross-document evidence integration (Khashabi et al., 2018; Welbl et al., 2018). Apparently complex QA questions may still be solvable through lexical, structural, or dataset-specific shortcuts, whereas genuine multi-hop questions require connected reasoning over distributed evidence (Chen \& Durrett, 2019; Trivedi et al., 2022).

In multilingual evaluation, comprehension depth may shape how the gap manifests. Cross-lingual QA performance can depend both on language-specific lexical and segmentation choices and on deeper semantic representations that transfer across languages (Artetxe et al., 2020; Vulić et al., 2020). However, cross-lingual performance gaps can vary nonmonotonically across question types rather than increasing uniformly with presumed cognitive demand (Liu et al., 2023; Yu et al., 2025). We therefore treat the \texttt{L0}--\texttt{L5} levels as an experimental factor rather than as a scale whose monotonicity is assumed.

QA responses are traditionally evaluated using exact-match and lexical-overlap metrics such as Exact Match and Token-F1 (Wang et al., 2017; Bulian et al., 2022). Lexical-overlap metrics are transparent and reproducible but can penalize semantically correct paraphrases, especially for open-ended answers (Bulian et al., 2022; Chen et al., 2019). BERTScore and entailment-based metrics aim to capture semantic equivalence beyond lexical overlap (Chen \& Eger, 2023; Kaster et al., 2021). LLM-based evaluators can exhibit self-preference, prompt and setup sensitivity, inconsistent judgments, and preferences for superficial stylistic features (Anghel et al., 2025; Chen et al., 2025; Jeong et al., 2025).

None of these metrics alone constitutes a direct measure of comprehension. Automatic metrics quantify observable properties of model outputs and are imperfect proxies for semantic adequacy or underlying comprehension (Sai et al., 2022; Mahowald et al., 2024; He et al., 2025). Substantive claims about model-output quality, including behavioral evidence of comprehension, should be validated against human judgments (Chiang \& Lee, 2023; Shen et al., 2023). In this study, the metrics score each response, whereas the executed primary CLCG is calculated as the difference between the English scored-row micro-mean and the pooled target-language scored-row micro-mean. Human validation is conducted separately to assess whether these automatic differences correspond to perceived differences in correctness, completeness, and semantic adequacy.

\subsubsection{Digital Resources, Tokenization, and Typology}\label{digital-resources-tokenization-and-typology}

Digital resource availability is one of the most frequently proposed explanations for unequal multilingual performance. Languages with more or higher-quality pretraining and instruction data tend to receive better linguistic coverage, stronger cross-lingual transfer, and better downstream performance during model development (Blevins \& Zettlemoyer, 2022; Li et al., 2024; Oladipo et al., 2023). However, speaker population and digital-resource availability are distinct dimensions of language resources: widely spoken languages may remain underrepresented in high-quality data, whereas languages with smaller speaker populations may have relatively strong digital infrastructure (Nigatu et al., 2024; Zaugg et al., 2022).

Tokenization is another important hypothesis. Tokenizers dominated by high-resource languages can fragment lower-resource languages more heavily, increasing token counts, inference cost, latency, and context-window consumption (Hong et al., 2024; Maksymenko \& Turuta, 2025). The effects of tokenization design and fragmentation on downstream performance, however, vary across model architectures, scripts, morphological profiles, vocabularies, and tasks (Clark et al., 2022; Hong et al., 2024; Qarah \& Alsanoosy, 2024).

Typological properties and writing systems have also been investigated as possible moderators of cross-lingual transfer. Linguistic distance, word order, morphological typology, language family, and writing system can influence cross-lingual transfer performance (de Vries et al., 2022; Martínez-García et al., 2021). Apparent typological effects can nevertheless be entangled with pretraining exposure, task-specific data availability, and tokenizer coverage (Park et al., 2021; Üstün et al., 2022).

This study tests digital resources, tokenization fertility, script, and typological status as candidate correlates of the CLCG. The objective is not to assume that any single factor explains the phenomenon, but to determine which associations remain consistent within the observed panel.

\subsubsection{Memorization, Prior Exposure, and New Content}\label{memorization-prior-exposure-and-new-content}

Language-model benchmarks built from static, publicly accessible content are vulnerable to training-data contamination or prior item exposure (Deng et al., 2024; Sánchez Salido et al., 2026). High performance may therefore reflect both the ability to solve the task and familiarity with the content or similar items. Because proprietary training corpora are not publicly auditable, external researchers generally cannot definitively verify whether specific benchmark items were included in training and must instead rely on imperfect indirect evidence (Deng et al., 2024; Mattern et al., 2023).

Declared knowledge cutoffs are coarse temporal indicators: they neither verify the inclusion or exclusion of specific documents nor guarantee that included content will be retrievable (Naser, 2025; Pericharla et al., 2026). Content published after a declared cutoff is less likely to have been part of the stated training period, but it may still have been incorporated through subsequent updates, external tools, or undocumented pipelines. Comparisons between content published before and after a cutoff should therefore be interpreted as sensitivity analyses rather than as proof that memorization is absent.

In this study, memorability is defined at the model--article level. An article is classified as potentially memorizable when its earliest availability date precedes the model's knowledge cutoff and as new when it follows that cutoff. The operational details of this classification and the corresponding sensitivity analyses are presented in the Methods and Supplementary Material. Within the related literature, this distinction establishes why prior exposure must be treated both as a threat to validity and as an analytical dimension.

\subsubsection{Positioning the CLCG}\label{positioning-the-clcg}

The CLCG is related to the multilingual evaluation literature, but it is not simply another benchmark score. Traditional metrics describe response quality under a single condition, whereas the CLCG describes how much of that quality is preserved when the language of the evidence changes while the other elements of the item remain constant. This distinction is methodologically similar to separating average performance from properties such as robustness, calibration, and fairness, whose evaluation requires explicit contrasts between conditions, distributions, or reference groups (Caton \& Haas, 2024; Höltgen \& Williamson, 2023; Ktena et al., 2024).

The CLCG is also not presented as a universal, instrument-independent property. Its magnitude depends on the task, domain, item type, model, reference condition, and evaluation protocol. The claim is narrower: under a controlled paired design, the CLCG provides a reproducible way to measure the relative preservation of contextual comprehension across languages.

The contribution of this study lies in combining a parallel \emph{within-item} experimental design with an executed aggregate micro-mean contrast, a panel that includes very-low- and ultra-low-resource languages, two question sources, human validation of the metrics, analyses of new content and stability, and reproduction in an external domain using FLORES-200. The next section formalizes the CLCG and defines the limits of its interpretation.

\subsection{The CLCG Construct}\label{the-clcg-construct}

\begin{figure}
\centering
\pandocbounded{\includegraphics[width=\linewidth,keepaspectratio]{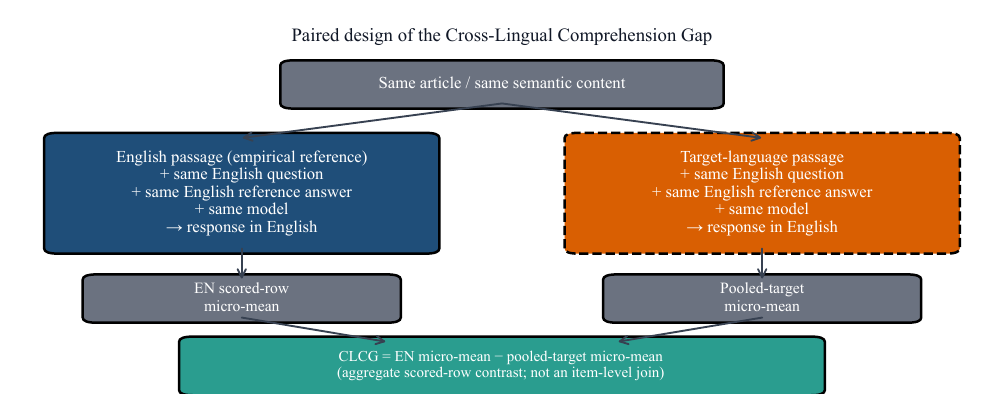}}
\caption{Paired design of the Cross-Lingual Comprehension Gap. Parallel design of the Cross-Lingual Comprehension Gap. The diagram shows the same model and item under aligned language conditions in which only the language of the evidence-bearing passage varies. The question, reference answer, model, and response language remain fixed. English serves as the empirical reference condition, and Portuguese is used separately as an empirical baseline for net CLCG. The figure depicts the design; the executed H1 estimator is the aggregate scored-row contrast defined in the Methods, not an item-level join.}
\label{fig:clcg-paired-design}
\end{figure}

We define the \emph{Cross-Lingual Comprehension Gap} (CLCG) as the systematic difference in a language model's semantic comprehension between a reference language and a target language when the task, item, and experimental controls are held constant and only the language of the text being read varies. Substantively, the CLCG captures how much of the ability to comprehend content is lost, or preserved, when the input language changes. It separates the cross-lingual contrast from other factors that also affect absolute scores, including item difficulty, the model's general capabilities, and characteristics of the reference answer. The parallel design is presented in Figure~\ref{fig:clcg-paired-design}, which identifies the components held constant and the component that varies between conditions. The diagram documents experimental parallelism at the level of the design; the executed estimator is defined and delimited in the Methods (Operational Definition of the CLCG).

The CLCG is a latent construct. It is not directly observable as a physical property of the model, nor is it equivalent to a raw leaderboard result. Like robustness, calibration, and fairness, a latent construct becomes measurable only through explicit operationalization. That operationalization must specify the estimator, empirical instrument, and interpretive controls (de Oliveira et al., 2025; Goyal et al., 2023; van der Wal et al., 2024). We therefore distinguish three layers. The \emph{construct} is the theoretical property, namely the cross-lingual comprehension gap. The \emph{estimator} is the procedure used to approximate it from observed scores. The \emph{ML metrics}, including Exact Match, Token-F1, NLI, and BERTScore, score the output within each condition. They are components of the estimator rather than the object of study.

This distinction addresses a foreseeable criticism that the CLCG is merely a new metric or a renaming of F1, NLI, or BERTScore. It is not. The CLCG does not replace F1, NLI, or BERTScore. It uses them to estimate a systematic cross-lingual difference in comprehension under controls that conventional performance benchmarks do not usually isolate. Conceptually, the estimator contrasts performance in the reference language with performance in the target language on the same items. The complete operational definition, including aggregation, depth bands, and extensions such as the net CLCG using Portuguese as a baseline, is presented in the Methods section.

The interpretation of the CLCG depends on its sign. A positive value indicates lower observed performance in the target-language condition, a value near zero indicates little observed difference, and a negative value indicates higher observed target-language performance. A near-zero estimate does not by itself establish statistical or practical equivalence. The same sign convention applies to net CLCG after replacing the English reference with the Portuguese baseline; formal definitions and worked examples are provided in Supplementary Material Section SM6.8.

We use English as a high-quality reference language. It is the language of both the gold answer and the model response in our design, and it functions as the source-language reference condition against which target-language performance is compared for the same item. This does not imply that English is a perfect semantic mirror of the passage in every target language. It means only that English provides the reference anchor against which the gap becomes comparable.

Finally, the construct validity of the CLCG can be examined through a nomological network by assessing whether the measure exhibits theoretically predicted relationships with resource availability, typological properties, and tokenization fragmentation or fertility (Bandarkar et al., 2024; Gerbing \& Anderson, 1988). The next section operationalizes the construct and describes the protocol used to estimate this difference and test these relationships.

\subsection{Methods}\label{methods}

This section describes the operationalization of the CLCG. The benchmark, corpus, and metrics are not the central objects of the study. Together, they form the empirical instrument used to observe the construct under controlled conditions. The design exposes the same models and items to parallel language conditions in which only the language of the evidence-bearing passage changes.

\subsubsection{Design Overview}\label{design-overview}

The operational scale of the study exceeds the size of the inferential sample. The reconstructed corpus contains 559 unique English source articles, 9,843 materialized article--language versions, 23,276 unique English body paragraphs, 410,186 aligned paragraph--language instances, and 23,091 unique questions, comprising 10,389 human-authored items and 12,702 LLM-generated items. The primary execution uses a stratified sample of 150 articles, containing 6,108 items, of which 6,026 are answerable under the scoring rules, evaluated across 18 languages and five models. Including two additional repetitions for the 40-article stability subsample, the execution plan comprised 844,920 potential response slots and produced 831,992 prediction records. Of these records, 826,308 were parse-valid. Excluding the 11,238 response rows associated with items for which \texttt{gold\_answerable=false} left 820,754 scored rows. Supplementary Material Section SM8 reports complete counts, denominator boundaries, the metric-level audit, and the structural-gap roster.

The study evaluates five language models from five distinct laboratories across a panel of 18 languages: English as the reference language, Portuguese as a high-resource baseline, and 16 target languages distributed along a gradient of digital-resource availability. The confirmatory sample contains 150 articles selected from the larger aligned panel of 559 article families, with structural alignment across the materialized language versions.

The primary task is contextual \emph{question answering}, organized into six levels of depth (\texttt{L0}--\texttt{L5}), ranging from polar verification and literal extraction to local inference, global synthesis, and interpretation of purpose. The instrument combines two item sources: questions generated under a controlled procedure by an external model and pre-existing human-authored questions from the corpus itself.

For each item, the model, question, reference answer, inference protocol, output language, and scoring procedure remain constant. Only the language of the passage read by the model varies. The questions, reference answers, and generated responses remain in English, reducing interference from target-language generation ability in the measurement of comprehension.

In the primary condition, the model receives the complete article body (\texttt{full\_body}) and answers all corresponding questions in a single call. This design preserves the information required for inference and synthesis items while amortizing the cost of the context across multiple questions.

\begin{figure}
\centering
\pandocbounded{\includegraphics[width=\linewidth,keepaspectratio]{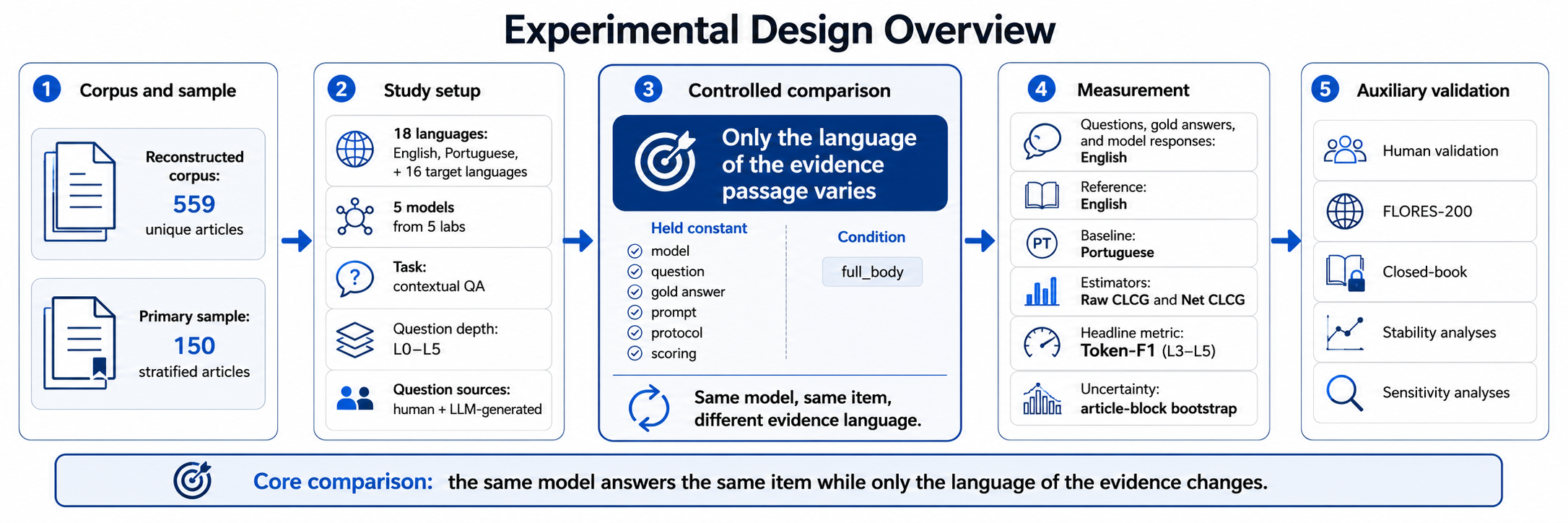}}
\caption{Experimental design overview. The reconstructed 559-article corpus supplies a stratified 150-article primary sample evaluated across 18 languages and five models. The primary controlled comparison holds model, item, question, gold answer, prompt, protocol, response language, and scoring constant while varying the evidence language under \texttt{full\_body}. Token-F1 in \texttt{L3}--\texttt{L5} supports raw and net CLCG estimates with article-cluster bootstrap; human validation, FLORES-200, \texttt{closed\_book}, stability, and sensitivity analyses remain auxiliary.}
\label{fig:experimental-design}
\end{figure}

Figure~\ref{fig:experimental-design} summarizes the executed experimental design and separates the primary CLCG pipeline from the auxiliary validation and robustness analyses.

The following subsections describe the parallel corpus and its translation quality, the construction of the comprehension instrument, the inference and scoring protocol, the statistical estimation of the CLCG, and the validation and robustness analyses.

\subsubsection{Models and Roles in the Pipeline}\label{models-and-roles-in-the-pipeline}

The study explicitly distinguishes evaluated models, which produce the responses used in the CLCG estimator, from auxiliary models, which are used to generate, verify, or analyze items. This separation prevents distinct functions from being treated as though they contribute equally to the primary measure.

The complete role matrix is provided in Supplementary Material Section SM2.1. It distinguishes the five evaluated models from four executed auxiliary-model roles, one deterministic validation mechanism, and an independent hybrid level-judge robustness audit (Supplementary Material Section SM11). Only responses from the five evaluated models enter the primary CLCG estimator. DeepSeek-V4-Flash appears in three auxiliary rows because the matrix is organized by pipeline role rather than by unique endpoint.

Item verification in the frozen pipeline relied on deterministic structural checks rather than a separate LLM-based level-verification pass. Generated items retained the level requested in the \texttt{L0}--\texttt{L5} generation call, and the stored field remains \texttt{level\_label\_validated=false}. The \texttt{\_validate\_item} function rejected malformed or off-contract items, checked \texttt{L1} spans against the passage, required valid supporting paragraph identifiers for \texttt{L3} and \texttt{L4}, and required at least two distinct identifiers for \texttt{L4}. An independent, generator-blind hybrid level judge---two \texttt{mistral-large-latest} verifiers emitting only JSON evidence, followed by a frozen deterministic map---was subsequently executed as a robustness audit of the depth labels on the full item bank (Supplementary Material Section SM11); the primary \texttt{L0}--\texttt{L5} field remains the generator design label and \texttt{level\_label\_validated} is unchanged. We additionally conducted a stratified manual validation of 300 question--answer items, comprising 50 items from each level.

The exact executed prompt texts, component hashes, output contracts, parsers, retry rules, and the hybrid level-judge robustness audit are documented in Supplementary Material Sections SM1 and SM11. Additional model-role and item-quality details are summarized in Supplementary Material Sections SM1 and SM2.

\subsubsection{Parallel Corpus and Translation Quality}\label{parallel-corpus-and-translation-quality}

A central threat to the construct validity of the CLCG is the possibility that the measured gap primarily reflects unequal source-translation quality rather than differences in model comprehension. We mitigate this threat through the properties of the instrument itself, rather than relying exclusively on statistical controls applied after data collection.

The textual material was collected from the Watchtower Online Library, or WOL, a public multilingual editorial source (Watchtower Online Library, n.d.). As part of the same multilingual publishing ecosystem, WOL provides access to Bible-based publications produced in more than 1,000 languages (Jehovah's Witnesses, 2022). This unusually broad coverage includes languages assigned to the lowest Joshi resource classes, which are commonly absent from multilingual benchmarks or represented only through machine translation. The present study does not use the complete collection. Instead, we selected an aligned parallel panel of 18 languages, comprising English as the reference language, Portuguese as a high-resource baseline, and 16 target languages for which the cross-lingual contrast was operationally feasible. The aligned panel contains 559 English source articles and 9,843 materialized article--language versions. All 559 articles are materialized in every panel language except Basque, which is available in the aligned corpus for 340 articles. Consequently, 340 article families contain all 18 languages, whereas 219 lack a Basque version. The metadata inventory lists 9,909 available article--language pairs, including 66 Basque entries that were not materialized in the aligned corpus; these metadata-only entries are not treated as aligned observations in the study.

The panel was selected to span a gradient of \emph{digital-resource availability}, the study's primary analytical dimension, while also diversifying language families and writing systems. Speaker counts and UNESCO endangerment labels in Table~\ref{tab:language-inventory} are dated Wikidata values for the panel languages and should not be conflated with digital-resource availability. Languages with large speaker populations may remain digitally underrepresented, whereas languages with fewer speakers may possess comparatively stronger digital infrastructure (Hammarström, 2015; Pimienta et al., 2023). Resource availability is coded using the external six-class taxonomy of Joshi et al.~(2020), from Class 0 (lowest availability) to Class 5 (highest availability). Ayacucho Quechua is left unclassified in the primary analysis because the published mapping does not unambiguously resolve that specific variety; assigning the generic Quechua entry to Class 1 is used only as a sensitivity analysis. Table~\ref{tab:language-inventory} summarizes the instrument.

{\footnotesize\setlength{\tabcolsep}{2.5pt}%
\begin{longtable}[]{@{}
  >{\raggedright\arraybackslash}p{(\linewidth - 14\tabcolsep - 1.25em) * \real{0.125}}
  >{\centering\arraybackslash}p{(\linewidth - 14\tabcolsep - 1.25em) * \real{0.05}}
  >{\raggedright\arraybackslash}p{(\linewidth - 14\tabcolsep - 1.25em) * \real{0.125}}
  >{\raggedright\arraybackslash}p{(\linewidth - 14\tabcolsep - 1.25em) * \real{0.125}}
  @{\hspace{1.25em}}
  >{\raggedright\arraybackslash}p{(\linewidth - 14\tabcolsep - 1.25em) * \real{0.20}}
  >{\centering\arraybackslash}p{(\linewidth - 14\tabcolsep - 1.25em) * \real{0.06}}
  >{\raggedright\arraybackslash}p{(\linewidth - 14\tabcolsep - 1.25em) * \real{0.18}}
  >{\raggedright\arraybackslash}p{(\linewidth - 14\tabcolsep - 1.25em) * \real{0.135}}@{}}
\caption[Languages included in the CLCG instrument.]{Languages included in the CLCG instrument. ISO code, Wikidata speaker count (with retrieval date), writing system, UNESCO endangerment risk (Risk), Joshi et al.~(2020) resource class, and role in the design. \textit{Note.} Data were retrieved from Wikidata on August 5, 2026. The query, retrieval code, and frozen dataset underlying this table are archived in da Silva \& Eicher (2026c).}\\
\toprule\noalign{}
Language & ISO & Speakers & Script & Risk & Joshi class & Class label & Role \\
\midrule\noalign{}
\endfirsthead
\toprule\noalign{}
Language & ISO & Speakers & Script & Risk & Joshi class & Class label & Role \\
\midrule\noalign{}
\endhead
\bottomrule\noalign{}
\endlastfoot
English & en & 1.13B\newline(2019/01) & Latin & --- & 5 & The Winners & Reference \\
Portuguese & pt & 254.30M\newline(2019/01) & Latin & 1 safe & 4 & The Underdogs & Baseline \\
Icelandic & is & 321.0k\newline(2015/01) & Latin & 1 safe & 2 & The Hopefuls & Target \\
Maltese & mt & 570.0k\newline(2012/01) & Latin & 1 safe & 2 & The Hopefuls & Target \\
Basque & eu & 750.0k\newline(2016/01) & Latin & 2 vulnerable & 4 & The Underdogs & Target (optional) \\
Georgian & ka & 3.70M\newline(2014/01) & Mkhedruli & 1 safe & 3 & The Rising Stars & Target \\
Haitian Creole & ht & 9.60M\newline(2007/01) & Latin & --- & 2 & The Hopefuls & Target \\
Swahili & sw & 82.30M\newline(2010/01) & Latin & --- & 2 & The Hopefuls & Target \\
Nepali & ne & 15.80M\newline(2019/01) & Devanagari & --- & 1 & The Scraping-Bys & Target \\
Cebuano & ceb & 15.90M\newline(2019/01) & Latin & 1 safe & 3 & The Rising Stars & Target \\
Xhosa & xh & 8.18M\newline(2011/01) & Latin & 1 safe & 2 & The Hopefuls & Target \\
Samoan & sm & 415.7k\newline(2015/01) & Latin & 1 safe & 1 & The Scraping-Bys & Target \\
Fon & fon & 1.94M\newline(2016/01) & Latin & --- & 0 & The Left-Behinds & Target \\
Guarani & gn & 4.85M\newline(1995/01) & Latin & --- & 1 & The Scraping-Bys & Target \\
Malagasy & mg & 18.14M\newline(2010/01) & Latin & --- & 1 & The Scraping-Bys & Target \\
Amharic & am & 21.90M\newline(2019/01) & Ge\textquotesingle{}ez & 1 safe & 2 & The Hopefuls & Target \\
Aymara & ay & 4.00M\newline(2020/01) & Latin & 2 vulnerable & 1 & The Scraping-Bys & Target \\
Ayacucho Quechua & quy & 918.2k\newline(2000/01) & Latin & 3 definitely\newline endangered & --- & Unresolved in primary mapping & Target \\
\end{longtable}
\label{tab:language-inventory}
}

The panel includes one language isolate, Basque, one quasi-isolate, Georgian, and three non-Latin writing systems: Devanagari, Ge\textquotesingle{}ez, and Mkhedruli. This diversity supports subsequent analyses of typology and tokenization fragmentation. Basque is an optional target in the unbalanced panel because its coverage is incomplete in part of the aligned corpus. The remaining languages exhibit stable alignment within the subset used in the study.

The suitability of this source rests on two complementary properties of its production. First, each publication is produced through a human, collaborative, multistage workflow that begins from an English source and coordinates translation across languages through shared source-text analysis, terminology tools, prior translations, and iterative review for meaning, naturalness, completeness, and correctness (``Breaking the Language Barrier,'' 2016). Second, much of this work is carried out through Remote Translation Offices, where teams work into their mother tongue from within or near the target-language community, helping maintain natural and locally current usage (\emph{2013 Yearbook of Jehovah's Witnesses}, 2013, pp.~26--28; Jehovah's Witnesses, 2021). This regional model is especially relevant for lower Joshi-class languages---Fon, Aymara, Ayacucho Quechua, and others underrepresented in multilingual benchmarks---which often lack parallel corpora, professional translation capacity, or reliable machine translation (Goyal et al., 2022; Ranathunga et al., 2023). Together these properties combine cross-version consistency with community-specific linguistic knowledge; they do not guarantee perfect semantic equivalence, but they reduce dependence on machine translation and the risk of unnatural text in less represented languages.

Under these conditions, a measured difference between English and a target language is interpretable, as a first-order approximation, as a difference in model comprehension rather than merely as an artifact of poor source translation. We nevertheless do not assume that translation is perfect. We report (i) automated completeness checks by language, (ii) translation spot checks for languages in the lowest represented resource classes, and (iii) a rule under which languages below the completeness threshold are treated as a stated limitation rather than as confirmatory evidence. The instrument therefore provides a professionally human-translated parallel corpus aligned at the paragraph level across a panel of 18 languages, with the documented Basque coverage limitation, and spans an explicit gradient of digital-resource availability. This combination offers a rare basis for observing the CLCG. The next subsection describes how the QA depth gradient was constructed over this aligned textual material.

\subsubsection{Construction of the Comprehension Instrument}\label{construction-of-the-comprehension-instrument}

The \emph{question-answering} task is not treated as the construct of this study. Instead, it serves as the instrument used to elicit and observe different forms of contextual comprehension. Rather than relying exclusively on literal or multiple-choice questions, we organize the items along a gradient of cognitive depth that ranges from verifying explicitly stated information to integrating information across the article.

The \texttt{L0}--\texttt{L5} gradient does not reproduce any single taxonomy or benchmark in full. It is a study-specific operationalization developed for the CLCG design by combining foundations from reading-comprehension research with established practices in \emph{machine reading comprehension}. The progression from literal comprehension to reorganization and inference is based primarily on Barrett's reading-comprehension taxonomy (Clymer, 1968). The design of \texttt{L5}, which targets interpretation of purpose and central argument, is informed by applications of the revised Bloom taxonomy to question design across cognitive levels (Crowe et al., 2008). The \texttt{L0} level, which includes the response \texttt{Not\ stated}, incorporates the principle of unanswerable items used in SQuAD 2.0 (Rajpurkar et al., 2018). The local-inference and global-synthesis levels also draw on research concerning multi-sentence and \emph{multi-hop question answering} (Yang et al., 2018). More broadly, the progression corresponds to the PISA reading processes of locating information, understanding through integration and inference generation, and evaluating and reflecting (OECD, 2019).

The external sources therefore ground the cognitive progression, but the operational definitions, response formats, and final mapping into six levels were adapted to the requirements of this study: maintaining a verifiable reference answer, controlling the output language, and distinguishing local extraction from integrative cross-lingual comprehension.

The six operational levels distinguish comprehension demand from expected response format and constitute a study-specific operationalization rather than a previously validated six-level psychometric scale. A consolidated definition of each level---the comprehension demand it targets, its grounding, the enforced structural criterion, and a worked example---is given in Supplementary Material Section SM1.3. Their construction and item-level verification are described in Supplementary Material Sections SM1.2--SM1.3 and SM2.4--SM2.5.

The levels represent cognitive depth rather than response length. We keep two dimensions of the instrument separate: comprehension demand and response format. This distinction is important because a longer response is not necessarily cognitively deeper, just as an inferential question may permit a short answer. Evaluation metrics are matched to the response format: Accuracy for categorical \texttt{L0} items, Exact Match and Token-F1 for constrained answers, and contextual semantic-similarity metrics such as BERTScore for open-ended responses (Clark et al., 2020; Zhang et al., 2020).

All automatically generated items were produced once from the English versions of the articles and subsequently frozen. Questions and reference answers remain in English under every condition. The same question is applied to the different language versions of the passage, ensuring that the task demand does not change across languages.

Using a fixed budget of four items per level per article (rather than scaling by text length), the pipeline produced a frozen bank of 12,702 LLM-generated items across the 559 articles (mean 22.72 per article; median 23; range 16--24), permitting fewer items when an article did not validly support a level rather than filling the quota. Per-article counts and the item-bank construction are detailed in Supplementary Material Sections SM1.2--SM1.3.

The automatic items were generated once by DeepSeek-V4-Flash and then frozen. This exact model version is not part of the evaluated panel. It nevertheless belongs to the same model family as DeepSeek-V4-Pro, one of the five evaluated models. We therefore do not claim complete model-family independence. We treat this proximity as a potential source of pro-DeepSeek bias and examine it through sensitivity analyses concerning item source and through comparison with pre-existing human-authored questions produced without the involvement of any LLM.

Each item records its generating model, prompt and template version, source article, requested level, and supporting sentences or paragraphs where applicable; the six per-level templates are in the reproducibility package and Supplementary Material Section SM1.2.

LLM-generated questions do not always reliably correspond to the requested level of cognitive demand and may concentrate at lower levels or differ from their intended categories (Law et al., 2025; Maity et al., 2025). An independent hybrid re-labeling (language-model evidence extraction followed by a deterministic structural map) was subsequently executed as a Supplementary Material robustness audit (Section SM11), and the depth-conditioned conclusions were preserved under it. Each item retained the level requested during generation after deterministic structural screening. Accordingly, the \texttt{L0}--\texttt{L5} field is treated as an intended design label rather than an independently validated cognitive classification, and analyses by level are interpreted with \texttt{level\_label\_validated=false}.

The generation contracts imposed anti-shortcut requirements on the inference levels (two-or-more-sentence integration for \texttt{L3}, at least two distinct supporting paragraphs for \texttt{L4}), enforced by deterministic structural screening; because the semantic booleans were not independently verified, these are prompt-level guardrails plus structural filters rather than independent validation of cognitive depth (Min et al., 2019; Trivedi et al., 2022). The full contracts and screening rules are in Supplementary Material Sections SM1.2--SM1.3.

All items must be answerable exclusively from the provided passage. The highest level of the gradient therefore represents integrative textual interpretation rather than subjective judgment, personal opinion, or external knowledge. This restriction preserves a verifiable reference answer and keeps the instrument focused on contextual comprehension.

In addition to the automatically generated items, the study uses pre-existing human-authored questions from the corpus itself. They provide a second item source independent of the generating model and serve as an ecological anchor for assessing whether the observed patterns persist in questions originally written for human readers. The next subsection describes this source and its integration into the instrument.

\subsubsection{Pre-existing Human-Authored Questions as an Ecological Anchor}\label{pre-existing-human-authored-questions-as-an-ecological-anchor}

In addition to the automatically generated items, the instrument incorporates 10,389 unique pre-existing comprehension questions from the corpus itself. The articles used in the study are organized into numbered paragraphs and include questions associated with specific parts of the text. Of the 23,276 English body paragraphs, 10,347 have one linked human-authored question and 12,929 have none; no English body paragraph has more than one question under the paragraph-linkage key used in the corpus audit. These questions were written by the articles' authors and editors for human readers as part of the publications' pedagogical method, before and independently of this experiment (Watch Tower Bible and Tract Society of Pennsylvania, n.d.).

This second item source serves a distinct methodological role. The LLM-generated items make it possible to control the distribution across the \texttt{L0}--\texttt{L5} levels and construct a depth-balanced design. The human-authored questions, in contrast, provide an ecological anchor: they represent questions that arose in an authentic reading and teaching context rather than questions produced specifically to maximize model performance or discrimination within a benchmark.

Each human-authored question is extracted together with its editorial link to the corresponding paragraph or set of paragraphs. This connection is preserved through structural fields in the source, including paragraph ordinal numbers and the identifiers used for alignment. These identifiers serve as provenance and alignment anchors: they locate the same structural evidence position across language versions without changing the content requested by the question. In the primary \texttt{full\_body} condition, however, the model receives the complete aligned article rather than only the linked paragraph or paragraphs.

\begin{figure}
\centering
\pandocbounded{\includegraphics[width=\linewidth,keepaspectratio]{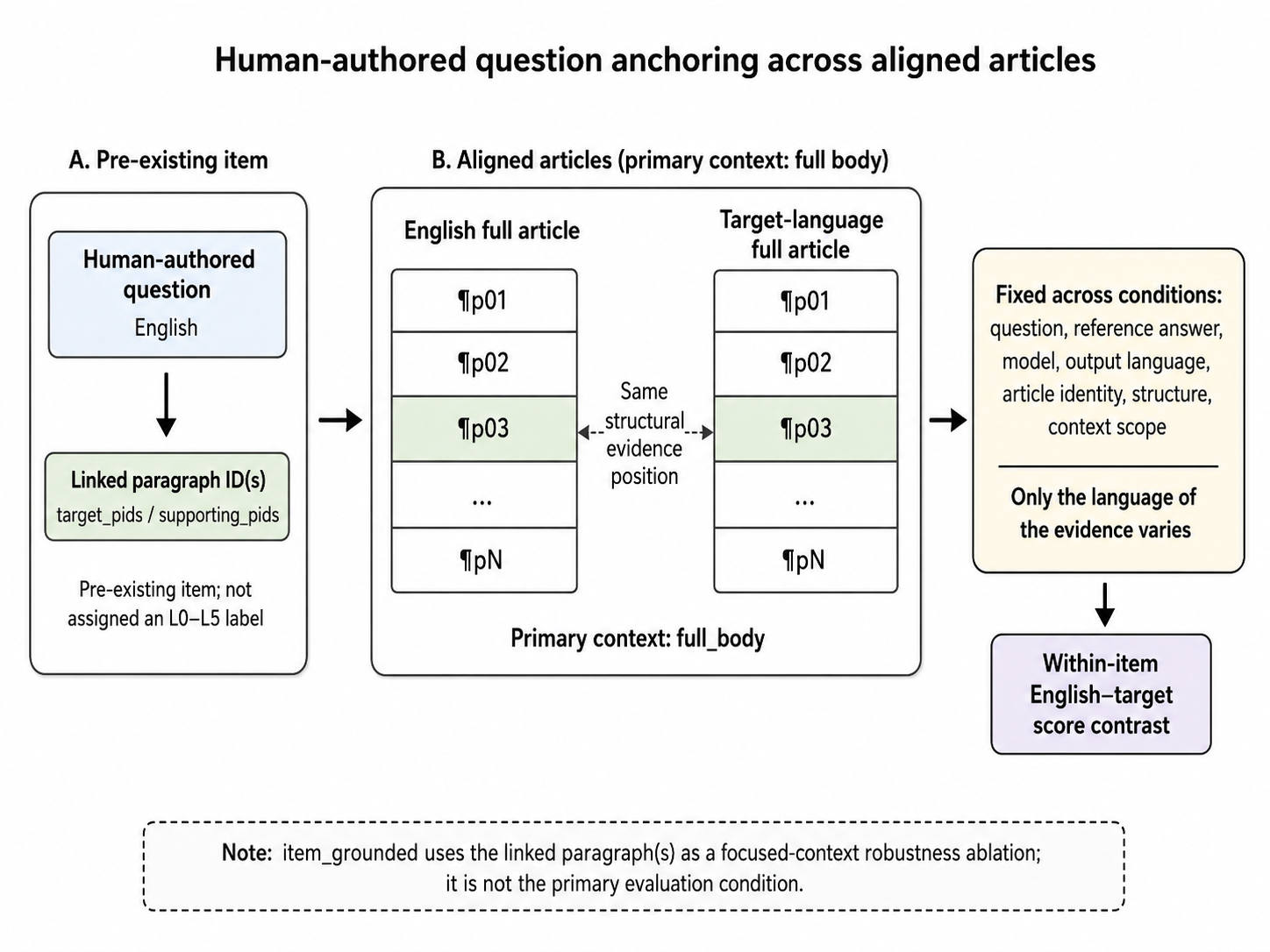}}
\caption{Human-authored question anchoring across aligned articles. A pre-existing English question retains its editorial link to one or more paragraph identifiers, which locate the corresponding evidence position across aligned article versions. In the primary \texttt{full\_body} condition, the complete article is supplied in each language while the question, English reference answer, evaluated model, output language, article identity, structure, and context scope remain fixed; only the language of the evidence varies. The linked identifiers define provenance and alignment rather than the sole primary context. The \texttt{item\_grounded} condition restricts context to the linked paragraphs; it is executed as a 40-article context-scope robustness probe (Supplementary Material Section SM10), separate from the primary \texttt{full\_body} analysis.}
\label{fig:human-question-alignment}
\end{figure}

Figure~\ref{fig:human-question-alignment} illustrates this mechanism for the human-authored item source. The question remains in English, the same structural evidence position is highlighted in the aligned English and target-language articles, and both conditions retain the complete article body. The question, English reference answer, evaluated model, output language, article identity, structure, and context scope remain fixed; only the language of the evidence varies.

Human-authored questions are retained as a separate category in the instrument rather than having their depth levels inferred and treated as equivalent to those of the generated items. This decision avoids introducing an additional classification as though it were directly observed. For descriptive purposes and secondary analyses, these questions may be mapped onto the \texttt{L0}--\texttt{L5} taxonomy through an independent procedure, but their primary role remains that of a distinct human-authored item source.

In the empirical design, the comparison between human-authored and LLM-generated items functions as a robustness test with respect to item source. If the direction and relative magnitude of the CLCG are preserved across both sources, the result becomes less dependent on choices specific to the generating model. This comparison is particularly relevant because DeepSeek-V4-Flash, which generated the automatic items, belongs to the same model family as one of the evaluated models.

The human-authored questions are not treated as a perfect psychometric gold standard. They were created for a specific pedagogical purpose, may focus primarily on local information, and were not originally balanced by depth or difficulty. Their methodological value lies in complementing the controlled instrument with a source external to the automatic generation process, rather than in replacing the experimental taxonomy.

\subsubsection{\texorpdfstring{Inference Protocol: Context Provided to the Model (\texttt{context\_scope})}{Inference Protocol: Context Provided to the Model (context\_scope)}}\label{inference-protocol-context-provided-to-the-model-context_scope}

A design decision in the inference stage determines the scope of evidence available to the model and, consequently, the form of comprehension assessed by the CLCG. Providing the complete article emphasizes long-context comprehension and information integration, whereas providing only the paragraph or paragraphs containing the answer emphasizes local extraction.

The primary condition is \texttt{full\_body}. The model receives the complete aligned article body in the target language, with paragraphs marked by \texttt{\P{}pid}, and answers all questions associated with that article in a single call for each article \ensuremath{\times} language \ensuremath{\times} model combination. The questions remain in English, and the answers are also produced in English under the \emph{output-in-English} control. This condition reproduces the context available to a human reader studying the article and is the only condition compatible with the integrative levels: \texttt{L4} multi-hop items require at least two paragraphs, whereas \texttt{L5} purpose questions require the article as a whole.

We do not use a single-paragraph condition as the general rule for four reasons. First, it is incompatible with the depth gradient. The LLM-generated questions were created from the complete article, and \texttt{L4}, \texttt{L5}, and some \texttt{L3} items cannot be answered from a single paragraph. Providing only one paragraph would therefore make these items unanswerable by design and would measure context removed by the experiment rather than comprehension. Second, it would create location leakage. Providing exactly the target paragraph would transform the task into short-span extraction or evaluation of a brief translated segment, thereby collapsing the intended construct. Third, it could change the construct being measured and potentially attenuate the observed gap. A short, location-revealing context removes much of the need to locate and integrate evidence across the article. The resulting task would therefore emphasize local decoding rather than the long-context comprehension targeted by the primary condition. Fourth, it would confound the comparison associated with H4. If paragraph-anchored human questions and multi-paragraph \texttt{L4} items received different amounts of context, the comparison between item sources would also become a comparison between context scopes.

The appropriate cost-control mechanism is therefore not to reduce the context but to use \emph{batching}. The article is provided once, and all questions are answered together, allowing the passage tokens to be shared across a mean of 40.72 items per article in the 150-article inference sample. The \texttt{full\_body} condition is expected to be more expensive than focused paragraph-level delivery because it processes a larger context. Batching nevertheless reduces repeated transmission of the same article across items while preserving substantially stronger construct validity.

Where focused context is methodologically appropriate, particularly for human-authored study questions and \texttt{L0}--\texttt{L2} items, the pipeline defines a versioned robustness condition called \texttt{item\_grounded}. Under this condition, each item receives only its \texttt{target\_pids} or \texttt{supporting\_pids}, typically one paragraph for human-authored and \texttt{L0}--\texttt{L2} items and at least two paragraphs for \texttt{L4}. The \texttt{item\_grounded} ablation is executed as a 40-article matched probe and reported in Supplementary Material Section SM10; on that panel the cross-lingual gap did not shrink under focused context ($\Delta=-0.028$, 95\% CI $[-0.037,-0.019]$).

In the executed study, \texttt{context\_scope=full\_body} is identical across languages for the same item. The English--target-language contrast therefore remains measurable. Context scope can change task difficulty and consequently the magnitude of the gap, which is why the \texttt{item\_grounded} context-scope probe (SM10) is kept distinct from the primary analysis. The following subsections describe the human validation of the metrics and the statistical protocol used for the estimator.

The inference prompt explicitly instructs the model to answer exclusively on the basis of the provided text, without relying on external knowledge, assumptions, or memorized information. When the required evidence is absent from the passage, the model must indicate that the answer is not supported by the text, following the response format defined for the item. This restriction serves as an attribution \emph{guardrail}. It discourages explicit reliance on parametric knowledge and keeps the task centered on contextual comprehension, but it cannot guarantee that the model will not recognize previously encountered content. That threat is addressed separately through the \texttt{closed-book} condition and the memorability analyses. Supplementary Material Section SM1 provides the executed templates, deterministic rules, parsing contracts, component hashes, and an annotated frozen execution example.

\subsubsection{Human Validation Study}\label{human-validation-study}

The evaluation framework and the CLCG estimator remain fully automated, reproducible, and scalable to any language available in the source. Human validation is conducted as a separate study whose primary purpose is to meta-evaluate the automatic metrics and whose secondary purpose is to compare the LLM-generated error taxonomy with human judgments. It is not part of the main measurement loop and is not required to run the instrument.

Because both the model output and the gold answer are always in English, annotators evaluate English against English. English proficiency is therefore sufficient, and speakers of the target languages are not required.

\paragraph{Recruitment, Eligibility, and Consent}\label{recruitment-eligibility-and-consent}

Annotators were recruited through the Clickworker crowdsourcing platform, targeting native English speakers in the United States, the United Kingdom, Ireland, Australia, or New Zealand; eligibility, self-reported consent, and the platform configuration are detailed in Supplementary Material Section SM9.1. The instrument comprised 666 unique evaluation items (600 individually evaluated responses and 66 contrastive pairs), each individual item assigned to three annotators under a load-balanced incomplete-block design.

The 600 individual responses were drawn from parseable \texttt{full\_body} predictions of the five evaluated models, stratified across 18 languages and three categories---\texttt{llm\_shallow} (\texttt{L0}--\texttt{L2}), \texttt{llm\_deep} (\texttt{L3}--\texttt{L5}), and \texttt{human}---with lower-quality responses oversampled to reduce ceiling concentration. Half of the sample used a hard quota of target-language responses carrying both automatic-judge (\texttt{DeepSeek}, \texttt{Mistral}) labels, supporting the human--automatic overlap; each response was still presented individually under the \emph{single} rubric. The full stratification and overlap construction are in Supplementary Material Section SM5.2.

Data collection ended after 200 annotators. A prespecified attention check embedded in the seventh item excluded 21 submissions in full, and two further annotators were excluded for pervasive internal contradictions; frozen item-level controls removed 42 additional annotator--item evaluations. The criteria were fixed before analysis and applied symmetrically, without reference to model, language, resource class, or automatic score. The final analytical sample retained 177 annotators and 1{,}727 valid item--annotator evaluations across 665 items (mean 2.597 valid annotators per item), yielding 10{,}407 dimension-level records, 138 decisive paired judgments, and 775 human--automatic overlap observations. Inter-rater agreement was computed with Krippendorff's \ensuremath{\alpha} on items with at least two valid ratings. The complete quality-control funnel, unit definitions, and agreement implementation are documented in Supplementary Material Sections SM5 and SM5.2.

Because adequacy is a global semantic judgment whereas the atomic-fact checklist measures exhaustive factual coverage, we retained in the primary analysis judgments in which maximum adequacy co-occurred with incomplete keypoint coverage, and conducted two post-hoc sensitivity analyses (removing the affected records, and removing every item with at least one such occurrence) that left the frozen rules, canonical ratings, and paired arm unchanged.

\subsubsection{Availability of Human Validation Data}\label{availability-of-human-validation-data}

As part of the study's public artifacts, we make the human-response export available through ParallelQA-18 after deidentification and the removal of confirmation codes, operational identifiers, and other potentially identifying fields. The repository contains the deidentified data, the analytical dataset resulting from the quality-control procedures, and the files required to reproduce the analyses (da Silva \& Eicher, 2026b). The transformations, exclusions, aggregations, and evaluation procedures are documented in the ParallelQA-18 Builder (da Silva \& Eicher, 2026a).

The public release does not include payment codes, Clickworker identifiers, email addresses, internal form links, or other metadata unnecessary for scientific reproducibility. Free-text responses that may have contained personal information were reviewed and redacted before publication. This organization makes it possible to audit the transition from the collected data to the analytical sample without exposing operational or personal information (Supplementary Material Section SM9.2).

\subsubsection{Human Rubric Anchored to the Automatic Metrics}\label{human-rubric-anchored-to-the-automatic-metrics}

The human rubric was designed to meta-evaluate specific components of the automatic pipeline rather than to produce a second, general, and independent measure of quality. Each field therefore has a predefined analytical purpose.

In the individual arm, annotators judge question validity, semantic adequacy, equivalence to the gold answer, attribution to the passage, and set-level consistency; when the reference answer contains multiple facts, a frozen atomic-fact checklist (generated once by DeepSeek-V4-Flash from the English question and gold answer, temperature zero, before data collection) yields \texttt{keypoint\_recall}. These fields map onto specific metric components---adequacy to NLI/BERTScore/Token-F1, equivalence to correspondence metrics, attribution to passage support, and the checklist to fluent-but-incomplete responses (Bulian et al., 2022; Chen \& Eger, 2023). A subsample additionally receives, last, the \texttt{primary\_error\_category} field, enabling comparison with the \texttt{clcg-error-v1} automatic judges without letting an error label bias the earlier judgments. In the contrastive arm, annotators choose between two anonymized responses to the same question from the same model (Answer A, Answer B, or a non-decisive option) with a confidence rating and randomized order; only A/B selections are decisive.

The complete executed rubric, exact response options and anchors, gold-fact decomposition, derived-variable rules, annotated example, and field-to-analysis mapping are documented in Supplementary Material Section SM5.2 and summarized in Supplementary Material Section SM5.

\subsubsection{Discriminative Power of the Human Validation Study}\label{discriminative-power-of-the-human-validation-study}

LLM-generated responses may appear fluent even when they omit important facts, attribute information incorrectly, or answer a question only partially. A human evaluation based solely on holistic impressions may therefore become concentrated near the top of the scale and lose the ability to distinguish substantively different responses.

Three mechanisms mitigate this risk: a sampling strategy that oversamples lower-quality responses within language/category/model cells so annotators see a strong--intermediate--weak mixture rather than near-perfect outputs; the atomic-fact checklist, whose \texttt{keypoint\_recall} records which gold-answer components were recovered and so keeps partially correct responses distinguishable from merely fluent ones; and the contrastive arm, which shows two same-model responses under different language conditions in randomized order and yields a directly interpretable \emph{win rate} that reduces dependence on absolute scales. The per-mechanism rationale is expanded in Supplementary Material Section SM5.7.

The contrastive arm was implemented within the final form allocation and yielded 138 decisive judgments after quality control. These judgments compare anonymized responses from the same model and item under higher-resource and lower-resource language conditions. The number actually analyzed is reported separately from the \emph{single-response} core, without treating each rubric dimension as an independent observation.

The primary purpose of human validation is not merely to demonstrate that the responses appear satisfactory. We seek two converging forms of evidence: first, that human evaluations track the automatic metrics; and second, that the cross-lingual gap is also reflected in human ratings, \texttt{keypoint\_recall}, and paired judgments.

We applied an executed ceiling rule that flags complete saturation only when the minimum adequacy median across represented resource classes is at least 4.5 and the mean adequacy difference between the HIGH and LOW automatic-score bins is below 0.5. No independent pre-analysis source available to the artifact audit jointly documented both thresholds and the compound rule, so we do not characterize this rule as preregistered. If the rule is triggered, \texttt{keypoint\_recall} and the contrastive \emph{win rate} receive greater interpretive weight. The complete accounting, rule, observed diagnostics, and post-hoc sensitivities are documented in Supplementary Material Section SM5.7 and summarized in Supplementary Material Section SM5.

\subsubsection{Auxiliary Error Analysis}\label{auxiliary-error-analysis}

In addition to estimating the magnitude of the CLCG, we conduct an auxiliary and descriptive analysis of how degradation manifests in model responses. This analysis remains outside the primary estimator, does not alter response scores, and does not support causal attribution of errors.

The frozen diagnostic sample contains 2,000 English--target-language response pairs. Each pair contributes one English baseline response and one target-language response from one of 16 non-English, non-Portuguese target-language strata. DeepSeek-V4-Flash and Mistral Small (\texttt{mistral-small-latest}) independently classify all 4,000 response sides, yielding 4,000 classifications per judge and 8,000 classifications in total. The builder omits explicit evaluated-model, ISO-code, and resource-class metadata, but this is metadata blinding rather than full language blinding because the passage itself remains visible and reveals or permits inference of its language.

Both judges apply the nine-label automatic fine taxonomy \texttt{clcg-error-v1}. The analysis additionally uses a seven-class collapsed automatic mapping, the separate seven-category human-facing rubric, binary semantic-error groupings, and a pair-level \texttt{crosslingual\_loss} condition. These are distinct analytical objects. The pair-level condition is satisfied when the English side is labeled \texttt{correct} and the target side receives one of the five executed semantic-error labels; it is not a tenth error label.

The sampling plan targeted 800 candidate-loss pairs (40\%), whereas the realized sample contains 921/2,000 (46.05\%). \texttt{candidate\_loss} is a sampling heuristic rather than a human or automatic error label. The diagnostic sample also uses seed 42 and a maximum of eight pairs per document.

The human overlap uses a design quota of 300 \texttt{ea\_overlap=true} target-language singles with dual automatic labels. The executed inventory contains 303 human-validation singles for which both labels are currently available: the 300-case quota core plus three spillover cases. After human quality control, the comparison contains 299 unique items and 775 individual annotator--item rows. Human--DeepSeek and human--Mistral agreement is computed on those individual rows; no per-item majority-vote human gold is implemented.

Selection of the 300-case core does not use quotas for DeepSeek--Mistral disagreement, non-\texttt{correct} labels, or individual error categories. Planned LOW/MID/HIGH score weights are 0.40/0.30/0.30, while the realized quota-core counts are 101/97/102. Within that core, 147/300 cases satisfy the \texttt{candidate\_loss} heuristic, both judges label 156/300 responses \texttt{correct}, and the judges disagree on 88/300 fine labels. These are emergent properties, not selection targets.

Automatic interjudge agreement is evidence of coding consistency rather than human validation or accuracy. Across 4,000 response-side comparisons, the judges agree exactly on the fine label in 3,057 cases, agree only after collapse in three cases, and remain unresolved in 940 cases. Human agreement with each judge is substantially lower, so neither automatic judge is treated as gold.

Supplementary Material Sections SM2.6--SM2.11 document the taxonomy definitions, mappings, prompt provenance, output schema, sample accounting, overlap reconciliation, and comparison units.

Error-analysis results are reported as stratified diagnostic frequencies and proportions. Because the protocol assigns one dominant primary category, they do not constitute an exhaustive multi-label decomposition or prevalence estimates for the complete scored dataset.

\subsubsection{Statistical Protocol}\label{statistical-protocol}

The scored response row is the basic observational unit, with grain \texttt{model\ \ensuremath{\times}\ language\ \ensuremath{\times}\ article\ \ensuremath{\times}\ repetition\ \ensuremath{\times}\ item}. The primary H1 code path does not create item-level English--target pairs. Instead, after restricting the score file to the \texttt{L3}--\texttt{L5} Token-F1 band, English, and the 16 target languages, it computes one micro-mean over English scored rows and one micro-mean over pooled target-language scored rows, then subtracts the latter from the former.

The executed H1 population contains 12,895 English rows and 202,776 target-language rows, for 215,671 scored rows in total. Each scored row has equal weight within its side, so languages with more available rows contribute more to the pooled target mean; English rows are not replicated once per target-language row. The parallel experimental design and estimator-level matching are therefore distinct properties.

The primary score is Token-F1, oriented so that higher values indicate better performance. Stage 5 retains scored parsing failures under the executed Rule A convention, assigning the fallback score already produced by the scorer; the primary H1 estimate is therefore not restricted to parse-valid rows. In the headline band, 214,172 rows are \texttt{ok}, 1,334 are \texttt{unparseable}, and 165 are \texttt{missing}. The primary failures-as-zero branch yields a pooled CLCG of 0.078. Excluding the 1,499 failures yields 0.076, a difference of 0.002.

All available repetitions enter the primary estimator. The 110 ordinary articles contribute one repetition, whereas the 40 stability articles contribute three. Consequently, stability articles contribute approximately three times the scored-row mass of ordinary articles before and within bootstrap resampling. The all-repetitions estimate is \texttt{0.07767880484501327}; restricting the analysis to \texttt{rep=0} yields \texttt{0.07946354774135533}, a difference of \texttt{\ensuremath{-}0.001784742896342062}.

Uncertainty is estimated through an article-cluster bootstrap. Each of 2,000 replicates samples 150 \texttt{doc\_id} clusters with replacement using seed 42. All rows associated with a selected article, including languages, models, items, and repetitions, retain the multiplicity of that article draw. The replicate recomputes English and target sums and counts and then the aggregate micro-mean contrast; it does not rebuild item-level pairs. The percentile interval uses the 2.5th and 97.5th percentiles and reproduces the frozen interval \texttt{{[}0.072,\ 0.084{]}}.

Supplementary Material Section SM3 documents parsing statuses and the alternative failure rule; Sections SM6.1--SM6.7 provide the estimator units, contrast contract, repetition weighting, bootstrap mechanics, and procedural pseudocode. These records describe the executed analysis but do not imply that every decision was preregistered.

\subsubsection{Operational Definition of the CLCG}\label{operational-definition-of-the-clcg}

Let \(r\) denote a scored response row and \(s_r\) its higher-is-better Token-F1 score. For the primary \texttt{L3}--\texttt{L5} analysis, let \(\mathcal{R}_{\mathrm{en}}\) contain the English rows and \(\mathcal{R}_{T}\) contain the pooled rows from the 16 target languages. The executed primary estimator is

\[
\widehat{\mathrm{CLCG}}_{\mathrm{micro}}
=
\frac{1}{n_{\mathrm{en}}}
\sum_{r\in\mathcal{R}_{\mathrm{en}}}s_r
-
\frac{1}{n_T}
\sum_{r\in\mathcal{R}_{T}}s_r,
\]

where \(n_{\mathrm{en}}=12{,}895\) and \(n_T=202{,}776\). This is an aggregate scored-row contrast, not the mean of item-level English--target differences. The executed H1 path does not construct an item-level inner join, and the English rows are not replicated once for each target-language row.

The pooled target mean can be decomposed as a coverage-weighted mean of the 16 language means. If \(n_l\) and \(\bar{s}_l\) denote the row count and score mean for language \(l\), then \(w_l=n_l/\sum_jn_j\) and

\[
\widehat{\mathrm{CLCG}}_{\mathrm{micro}}
=
\sum_{l=1}^{16}
w_l\left(\bar{s}_{\mathrm{en}}-\bar{s}_l\right).
\]

Thus, the primary estimator gives more weight to languages with more available scored rows. Its exact reconstruction is \texttt{0.07767880484501327}, reported as \texttt{0.078} with a 95\% article-cluster bootstrap interval of \texttt{{[}0.072,\ 0.084{]}}.

For descriptive language-level reporting,

\[
\widehat{\mathrm{CLCG}}_l
=
\bar{s}_{\mathrm{en}}-\bar{s}_l,
\]

and the equal-language macro summary is

\[
\widehat{\mathrm{CLCG}}_{\mathrm{macro}}
=
\frac{1}{16}
\sum_{l=1}^{16}
\widehat{\mathrm{CLCG}}_l.
\]

The exact macro estimate is \texttt{0.0774765487938531}, displayed as \texttt{0.077}. The micro and macro estimates differ because the former weights scored rows and the latter assigns weight \(1/16\) to every target language. Their exact difference is \texttt{0.0002022560511601712}; the difference \texttt{0.001} obtained from the displayed values is only a rounded comparison.

Portuguese is an empirical baseline, not a neutral condition or multilingual gold standard. Its raw gap is

\[
\widehat{\mathrm{CLCG}}_{\mathrm{pt}}
=
\bar{s}_{\mathrm{en}}-\bar{s}_{\mathrm{pt}},
\]

with exact value \texttt{0.0610478092283831}, displayed as \texttt{0.061} with interval \texttt{{[}0.055,\ 0.067{]}}. For language \(l\), the net gap is

\[
\widehat{\mathrm{CLCG}}^{\mathrm{net}}_l
=
\widehat{\mathrm{CLCG}}_l
-
\widehat{\mathrm{CLCG}}_{\mathrm{pt}}
=
\bar{s}_{\mathrm{pt}}-\bar{s}_l.
\]

The equal-language macro net estimate is

\[
\widehat{\mathrm{CLCG}}^{\mathrm{net}}_{\mathrm{macro}}
=
\widehat{\mathrm{CLCG}}_{\mathrm{macro}}
-
\widehat{\mathrm{CLCG}}_{\mathrm{pt}},
\]

with exact reconstruction \texttt{0.016428739565470002}, displayed as \texttt{0.016} with interval \texttt{{[}0.013,\ 0.020{]}}. This value is a macro net summary. The optional row-weighted micro net quantity is instead \texttt{0.016630995616630173}, which rounds to \texttt{0.017} when calculated from the displayed micro H1 and Portuguese gap. The two estimands must not be conflated.

A difference of two means can be rewritten as a mean of numerical differences for equally sized sequences with the same normalized weights. Semantic alignment is not required for that arithmetic identity, but it is required to interpret each difference as a meaningful within-unit paired contrast. The executed H1 satisfies neither the equal-denominator condition for a pooled pairwise rewrite nor the scientific matching condition: it has different side denominators and does not construct matched English--target units.

Raw and Portuguese-baseline quantities are combined within each bootstrap replicate when constructing the net interval. Confidence-interval endpoints are not subtracted, because doing so would discard the covariance between the two components.

Supplementary Material Section SM6.8 provides the notation, 16-language weighting decomposition, pairwise special-case boundary, exact micro and macro net reconstructions, bootstrap-net identity, worked examples, and numerical audit. Sections SM6.1--SM6.7 govern execution and bootstrap mechanics. The Methods define sign interpretation, while Sections SM6.9--SM6.11 report hypothesis tests and equivalence criteria.

\subsubsection{Confirmatory Inference}\label{confirmatory-inference}

The study tests four confirmatory hypotheses, each written against a null of no cross-lingual effect. \textbf{H1} asks whether a gap exists at all: the pooled English score exceeds the pooled target-language score (null: the two are equal). \textbf{H2} asks whether potentially memorizable content narrows the gap relative to novel content (null: no difference). \textbf{H3} asks whether the gap widens with comprehension depth from \texttt{L0} to \texttt{L5} (null: a flat profile). \textbf{H4} asks whether the gap depends on how items were written, comparing LLM-generated with human-authored questions (null: no difference). Throughout, we treat a hypothesis as supported when its article-cluster bootstrap interval excludes the null and the result survives the Benjamini--Hochberg correction described below; the mixed-effects models serve only as convergent checks and never as sole support.

Confirmatory tests are evaluated using two complementary strategies. The primary strategy uses an article-level cluster bootstrap with 2,000 resamples. In each resample, all scored rows associated with the same article remain grouped together. For H1, the replicate recomputes English and pooled-target sums and counts and then their micro-mean contrast; it does not reconstruct item-level pairs.

As a convergent analysis, we fit mixed-effects models with crossed random effects for article and model. Results from this second strategy are reported alongside the bootstrap results but are not used alone to support confirmatory conclusions. Agreement between an identifiable, converged mixed model and the bootstrap reduces dependence on a single inferential specification; a nonconverged model is reported as such and provides no independent support.

The executed false-discovery-rate family contains ten tests: the aggregate H1 CLCG, five model-specific H1 contrasts, the H2 memorability bootstrap contrast, the H2 mixed-model interaction, the H3 standardized item-depth slope, and the H4 item-source difference. We control the false discovery rate across these entries using the Benjamini--Hochberg procedure (Benjamini \& Hochberg, 1995). Both raw and adjusted \(p\)-values are reported. The H4 and Portuguese equivalence diagnostics are reported separately from this multiplicity family.

Directional hypotheses encoded in the executed analysis, such as a CLCG greater than zero, are evaluated using the bootstrap distribution of the contrast in the predicted direction. Confidence intervals remain the primary representation of uncertainty, and statistical significance does not replace interpretation of effect magnitude.

Portuguese is evaluated as an empirical high-resource baseline rather than as a condition whose expected gap is necessarily zero. To determine whether the English--Portuguese contrast is sufficiently small to be considered practically equivalent for the stated analytical purpose, we use the two one-sided tests procedure with the executed equivalence bounds (Lakens, 2017; Schuirmann, 1987). Failure to establish equivalence does not invalidate the baseline. Instead, it indicates a measurable cost of operating outside English, which is precisely the component subtracted by the net CLCG.

Testing the item-depth gradient requires caution because the levels use different response formats and metrics. Direct comparisons of raw scores across \texttt{L0}--\texttt{L5} could conflate cognitive depth with metric scale. The confirmatory analysis therefore standardizes scores within each level and estimates the association between depth and the gap. Raw level-specific results remain descriptive.

The memorability analysis is conducted only when the evaluated set contains sufficient variation between items published before and after each model's knowledge cutoff. When the factor is constant or insufficiently represented, the test is omitted and its non-identifiability is reported explicitly. DeepSeek-V4-Pro has no novel article in the confirmatory band and is therefore omitted from its within-model contrast. The pooled H2 mixed-effects model did not converge and is not treated as independent confirmatory support. The \texttt{closed-book} control described below provides a complementary experimental check by measuring how much performance remains when the passage is removed.

The hypotheses, equivalence bounds, multiplicity family, mixed-model specifications, and rules for non-identifiable tests are reported in Supplementary Material Sections SM6.9--SM6.11. These records document the executed analysis but do not retrospectively establish preregistration. Inferential findings are interpreted in the Results section.

\subsubsection{Nomological Network of the CLCG}\label{nomological-network-of-the-clcg}

In addition to the confirmatory hypotheses, we explore whether the CLCG is associated with external characteristics that should theoretically influence cross-lingual performance. This analysis examines whether the construct behaves consistently with expectations concerning language-resource availability, tokenizer fragmentation, writing system, and a coarse typological probe. It remains exploratory and outside the confirmatory family defined in Supplementary Material Sections SM6.9--SM6.11.

The primary external variable is the level of digital resources. Languages are assigned to the six-class taxonomy proposed by Joshi et al.~(2020), in which Class 0 denotes the lowest and Class 5 the highest resource availability. The primary analysis uses the published mapping, excludes Ayacucho Quechua because its specific variety could not be resolved unambiguously, and treats the generic Quechua mapping to Class 1 only as a sensitivity analysis. We expect larger gaps at lower class values, but interpret the association as a general tendency rather than a deterministic relationship.

The executed continuous proxy is English-centric tokenizer fertility. Using \texttt{tiktoken} with the \texttt{cl100k\_base} encoding, we calculate the mean ratio of tokens in each language to tokens in English over parallel non-short body paragraphs from a seeded sample of 40 aligned articles. This operationalization is tokenizer- and sample-specific; it is not treated as a universal property of a language or of every evaluated model's tokenizer.

The executed linguistic probes are binary indicators for non-Latin script and isolate or quasi-isolate status. The latter identifies Basque and Georgian from the deployed language-family metadata. These coarse indicators do not constitute complete measures of typological, structural, lexical, or phonetic distance.

Writing system receives separate attention because tokenization and segmentation may affect token counts, context costs, and internal representations. We examine its bivariate association with the CLCG and a partial Spearman sensitivity that adjusts approximately for canonical Joshi class.

The primary exploratory summaries are language-level Spearman correlations across Portuguese and the 15 target languages with canonical Joshi classes. Their 95\% intervals use 2,000 percentile resamples of the 16 analyzed languages with seed 42. We additionally report rank-residual partial Spearman analyses for fertility and non-Latin script controlling Joshi class; no confidence intervals were stored for these partial analyses. A standardized three-predictor OLS is retained as a descriptive multivariable diagnostic without standard errors, confidence intervals, or inferential \(p\)-values.

Ayacucho Quechua remains outside the primary factor analysis because its variety has no unambiguous canonical Joshi mapping. A separate sensitivity assigns the broader Quechua Class 1 proxy, increasing the language-level sample from 16 to 17 without changing its canonical classification. Model-by-factor interactions and alternative linguistic-distance analyses were not executed and are not represented as results.

These analyses are not used to redefine the CLCG or support causal claims. Because the language-level sample is small and the predictors are correlated, results are interpreted in terms of magnitude, direction, uncertainty, and stability rather than as independently identified effects.

Supplementary Material Sections SM6.12--SM6.14 provide the complete 18-language inventory, executed codings, formulas, predictor-dependence matrix, missingness rules, row-level mapping, and sensitivity boundaries.

\subsubsection{\texorpdfstring{\texttt{Closed-Book} Control and Context Gain}{Closed-Book Control and Context Gain}}\label{closed-book-control-and-context-gain}

Instructing models to answer only on the basis of the passage reduces explicit reliance on external knowledge, but it does not guarantee that the response was derived from the provided context. Models may recognize previously encountered content, retrieve parametric knowledge, or exploit cues contained in the question itself. We therefore complement the primary protocol with a \texttt{closed\_book} condition.

Under the primary \texttt{full\_body} condition, the model receives the complete article in the evaluated language and answers the questions in English. Under the \texttt{closed\_book} condition, no passage is supplied and the model receives the same English questions. For matched rows, model identity, item IDs, questions, gold answers, English response language, JSON-array output structure, parser family, and native scoring rule are aligned. The prompts and generation settings are not identical: \texttt{full\_body} instructs passage-only answering and uses \texttt{temperature\ =\ 0.2} with seed \texttt{42}, whereas \texttt{closed\_book} instructs the model to answer from prior knowledge and best judgment and uses \texttt{temperature\ =\ 0.0} with seed \texttt{20260322}. Both request \texttt{top\_p\ =\ 1.0} and reuse model-specific maximum-token tuning where applicable. Passage availability is therefore the principal experimental contrast, but not the only procedural difference.

We define context gain as the difference between the score obtained with the passage and the score obtained without it:

\[
G^{\mathrm{context}}_{m,i,l}
=
s^{\mathrm{full\_body}}_{m,i,l}
-
s^{\mathrm{closed\_book}}_{m,i}.
\]

Positive values indicate that access to the passage improves performance. Values near zero indicate that the model performs similarly without access to the text, whereas negative values may reflect generation variability, contextual interference, or scoring instability.

Performance under the \texttt{closed\_book} condition is not interpreted as direct evidence of verbatim memorization. A correct answer without context may result from general knowledge, prior exposure to the content, cues in the question, or coincidence. The condition instead serves as a diagnostic of answerability without the passage and of the model's effective reliance on the provided context.

This control also distinguishes between two phenomena. A model may have a low absolute score in a language while still obtaining a large gain from the passage, suggesting that the text contributes useful information even when comprehension remains limited. Conversely, similar performance with and without context provides little evidence that the answer resulted from contextual reading, even when the final response is correct.

Cross-language comparisons use paired context gains for the same model and item. We examine whether lower-resource languages yield smaller gains than English or Portuguese, which would be consistent with a reduced ability to transform the provided evidence into a correct response. This analysis complements the CLCG. The CLCG measures performance differences across languages when context is available, whereas context gain measures how much the presence of the passage contributes within each condition.

The executed selector takes the first six numerically sorted document IDs found in the \texttt{gpt-5.6-terra} prediction directory and, within each document, the first two lexicographically sorted item IDs at each of \texttt{L3}, \texttt{L4}, and \texttt{L5}. This yields 36 items. The selector does not inspect knowledge cutoffs, the \texttt{memorizable} field, or observed \texttt{full\_body} performance, and the sample is not treated as a memorability-prioritized or performance-prioritized sample.

Supplementary Material Section SM4 provides the exact 36-item roster, model-presence flags, prompt and parameter comparison, pairing rules, frozen results, result mapping, and inferential boundaries. Results are reported separately from the primary estimator and are not used to automatically exclude items from the CLCG calculation.

\subsection{Results}\label{results}

\subsubsection{Variation Across Languages and Resource Classes}\label{variation-across-languages-and-resource-classes}

The results are organized as a single argument: we first establish the gap, then show that it survives every attempt to explain it away---alternative metrics, repeated decoding, likely prior exposure, a narrowed context scope, and an entirely different corpus---and finally that human raters perceive it too.

In the primary \texttt{L3}--\texttt{L5} condition, the English scored-row Token-F1 mean was 0.456. The executed pooled scored-row contrast produced a CLCG of 0.078 (95\% CI: 0.072--0.084), corresponding to an approximately 17\% reduction relative to the English reference score. The equal-language macro summary across the 16 target languages was 0.077. The proportional value is reported only as a descriptive aid; the primary estimand is the absolute aggregate English--target micro-mean contrast.

The aggregate CLCG masks substantial heterogeneity across the evaluated languages. Although most target languages exhibited lower performance than English, the magnitude of the gap varied by language and digital-resource level.

Under the canonical Joshi et al.~(2020) taxonomy, the gap broadly widened as resource availability fell. The mean raw CLCG was 0.063 in Class 4, 0.070 in Class 3, 0.065 in Class 2, 0.085 in Class 1 (The Scraping-Bys), and 0.104 in Class 0 (The Left-Behinds); the corresponding net gap over Portuguese reached 0.043 in Class 0. Class 0 contained only Fon and is therefore a descriptive singleton rather than a generalizable group estimate, and the Class 2--3 ordering is mildly non-monotonic given the small, uneven class sizes. No analyzed target language belonged to Class 5, and Ayacucho Quechua was excluded from primary class aggregates because its mapping was unresolved. Full per-class intervals are given in Supplementary Material Sections SM7.1--SM7.2.

The equal-language macro estimate is \texttt{0.077}, compared with the primary row-weighted H1 estimate of \texttt{0.078}. The class analysis includes Portuguese and excludes unclassified Ayacucho Quechua. Complete language-level and resource-class results are reported in Supplementary Material Sections SM7.1--SM7.2, and estimator details are provided in Sections SM6.1--SM6.7.

\begin{figure}
\centering
\pandocbounded{\includegraphics[width=\linewidth,keepaspectratio]{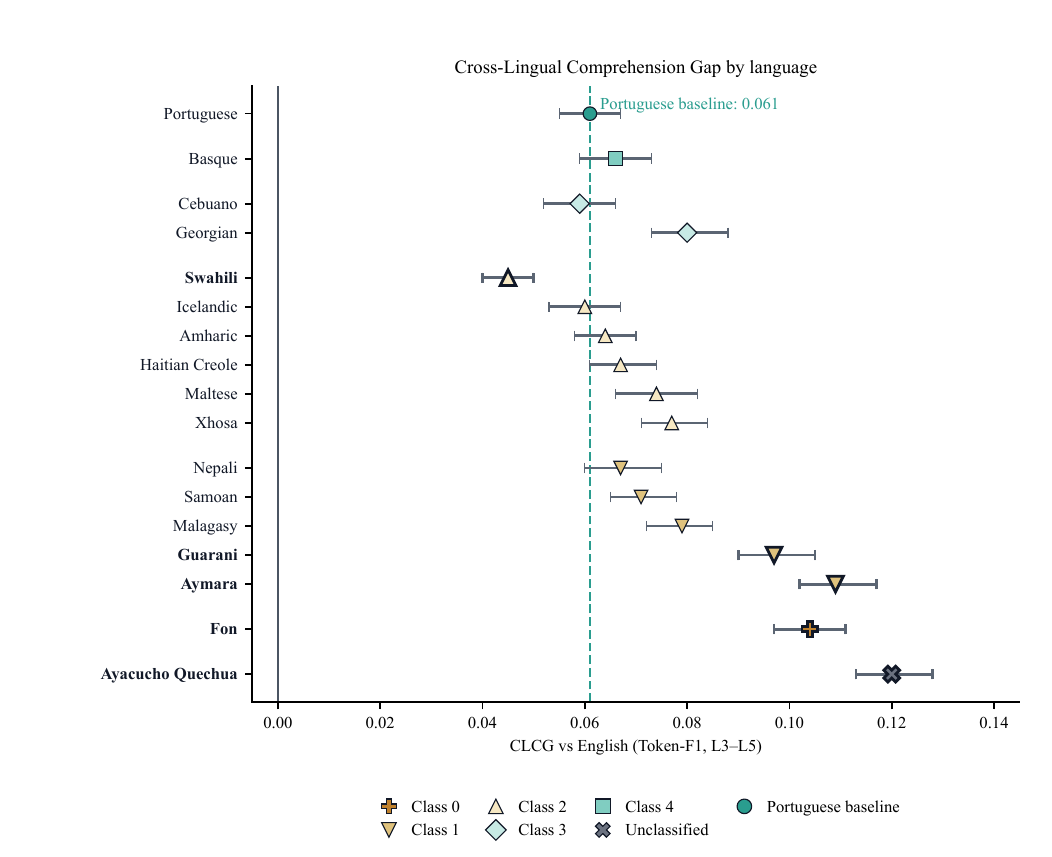}}
\caption{Cross-Lingual Comprehension Gap by language. Points show the language-level raw CLCG, defined as the English Token-F1 score minus the score obtained when the same evidence was presented in each language, in the primary \texttt{L3}--\texttt{L5} band and pooled across the five evaluated models. Horizontal bars represent 95\% percentile confidence intervals obtained from 2,000 bootstrap resamples clustered by article. The solid vertical line marks zero, corresponding to the English reference condition, and the dashed green line marks the Portuguese baseline gap of 0.061. Resource classes follow Joshi et al.~(2020); Ayacucho Quechua remains unclassified in the primary mapping, and no analyzed target language belongs to Class 5. Higher values indicate lower relative preservation. Bold labels identify languages highlighted in the accompanying discussion and do not denote statistical significance. The grouping is descriptive and does not imply a causal effect of resource availability.}
\label{fig:clcg-by-language}
\end{figure}

To see whether the aggregate result is driven by a few extreme cases, Figure~\ref{fig:clcg-by-language} plots the raw CLCG for Portuguese and each of the 16 target languages, with article-cluster bootstrap intervals. Organizing languages by Joshi resource class exposes both within-class dispersion and between-class differences.

\begin{figure}
\centering
\pandocbounded{\includegraphics[width=\linewidth,keepaspectratio]{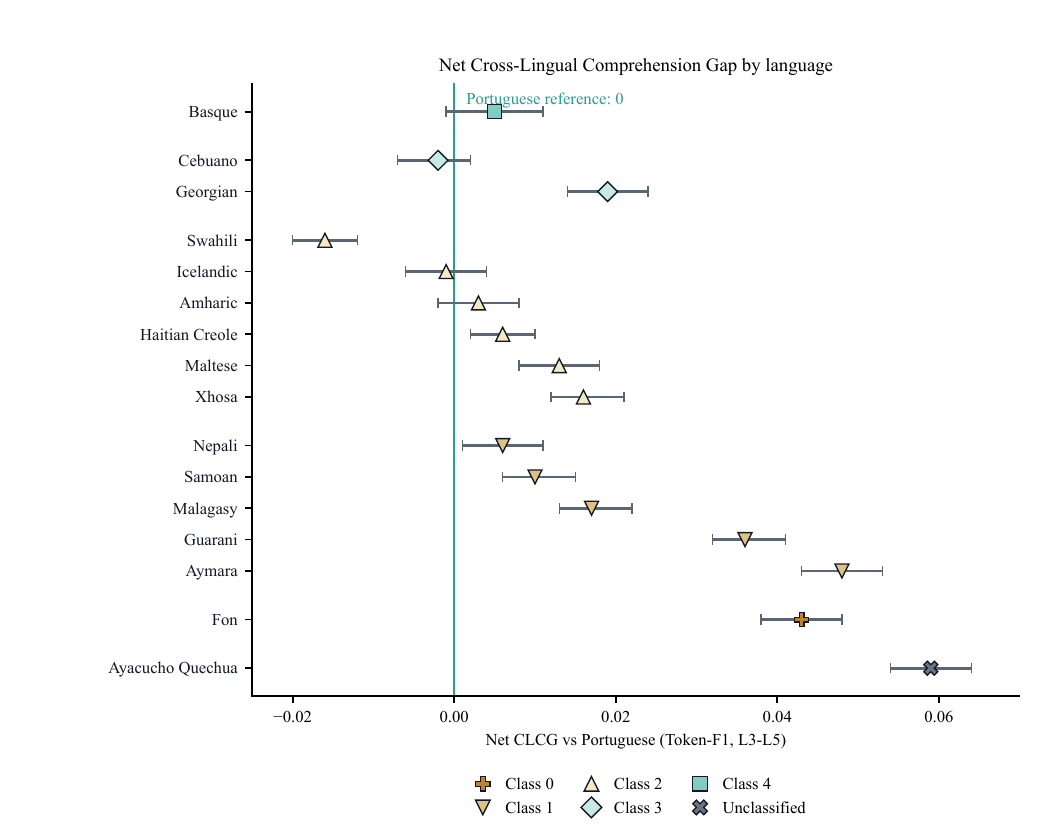}}
\caption{Net Cross-Lingual Comprehension Gap by language. Points show the stored language-level net CLCG, defined as Portuguese Token-F1 minus target-language Token-F1, in the primary \texttt{L3}--\texttt{L5} band and pooled across the five evaluated models. Horizontal bars are the stored 95\% percentile confidence intervals from 2,000 bootstrap resamples clustered by article. The vertical line at zero marks the Portuguese reference; Portuguese and English are therefore not plotted as result rows. Positive values indicate lower observed Token-F1 than Portuguese, while negative values indicate higher observed Token-F1. Resource classes follow Joshi et al.~(2020), with Ayacucho Quechua retained as unclassified. Class encoding is descriptive and does not imply causation. Estimates and raw components are rounded independently; the resulting 0.001 display residual for Malagasy does not alter the stored net estimate or its interval.}
\label{fig:net-clcg-by-language}
\end{figure}

Figure~\ref{fig:net-clcg-by-language} isolates the portion of each language-level gap that remains relative to the empirical Portuguese baseline. The stored net estimates are shown directly rather than reconstructed from rounded raw components. Net losses were largest for Ayacucho Quechua (0.059, 95\% CI: {[}0.054, 0.064{]}), Aymara (0.048, {[}0.043, 0.053{]}), and Fon (0.043, {[}0.038, 0.048{]}). Swahili instead had a negative net estimate (-0.016, {[}-0.020, -0.012{]}), indicating higher observed Token-F1 than Portuguese in this band. Several other intervals crossed zero and are not interpreted as evidence of equivalence.

Class comparisons are treated as descriptive and exploratory analyses. Digital-resource level is an aggregate characteristic correlated with other factors, including coverage in training corpora, benchmark availability, language family, writing system, and tokenization efficiency. Differences between classes are therefore not interpreted causally.

Nevertheless, the observed pattern is consistent with the central hypothesis of the study: the ability to retrieve and integrate information from a parallel passage deteriorates more sharply in languages that are less represented in the digital ecosystem and in language-model development.

\subsubsection{Variation Across Models and Comprehension Levels}\label{variation-across-models-and-comprehension-levels}

\begin{figure}
\centering
\pandocbounded{\includegraphics[width=\linewidth,keepaspectratio]{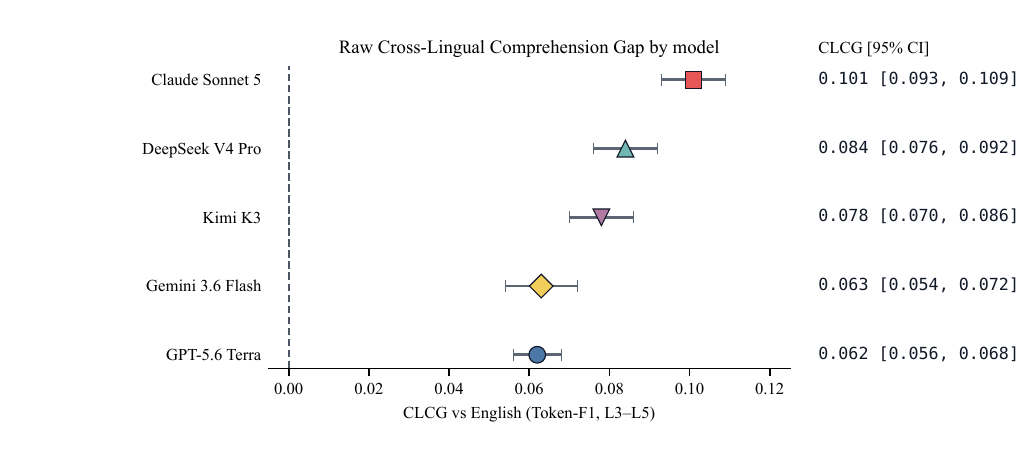}}
\caption{Raw Cross-Lingual Comprehension Gap by model. Points show the model-level raw CLCG, defined as the difference between English Token-F1 and target-language Token-F1 in the primary \texttt{L3}--\texttt{L5} band. Horizontal bars represent 95\% percentile confidence intervals obtained from 2,000 bootstrap resamples clustered by article. Models are ordered by decreasing CLCG. All intervals remain above zero, indicating that the cross-lingual loss was observed for each evaluated model. Lower CLCG indicates greater relative preservation across languages and should not be interpreted as higher overall model quality.}
\label{fig:clcg-by-model}
\end{figure}

Figure~\ref{fig:clcg-by-model} shows that the CLCG was positive for all five evaluated models, although its magnitude varied substantially across them. In the primary condition, comprising \texttt{L3}--\texttt{L5} items, GPT-5.6 Terra exhibited the smallest raw gap, with a CLCG of 0.062 and a 95\% confidence interval of 0.056 to 0.068. Gemini 3.6 Flash produced a similar estimate of 0.063 {[}0.054, 0.072{]}. These were followed by Kimi K3 at 0.078 {[}0.070, 0.086{]} and DeepSeek V4 Pro at 0.084 {[}0.076, 0.092{]}. Claude Sonnet 5 exhibited the largest gap, at 0.101 {[}0.093, 0.109{]}.

All confidence intervals remained entirely above zero, indicating that the phenomenon was not driven by a single model. At the same time, the range from 0.062 to 0.101 shows that the models differ in how well they preserve comprehension when the language of the evidence changes.

The ranking based on absolute English performance did not directly correspond to the ranking based on the CLCG. Claude Sonnet 5 achieved the highest English score, 0.489, but also exhibited the largest cross-lingual gap. In contrast, GPT-5.6 Terra and Gemini 3.6 Flash achieved lower absolute English scores but smaller losses between English and the target languages. Absolute English performance and cross-lingual robustness are therefore related but non-equivalent dimensions.

Decomposition using the Portuguese baseline confirmed that the result was not limited to a general cost of operating outside English. The net CLCG was 0.022 for Claude Sonnet 5 {[}0.017, 0.027{]}, 0.022 for DeepSeek V4 Pro {[}0.014, 0.029{]}, 0.009 for Gemini 3.6 Flash {[}0.003, 0.014{]}, 0.012 for GPT-5.6 Terra {[}0.008, 0.016{]}, and 0.019 for Kimi K3 {[}0.014, 0.023{]}. All intervals excluded zero, indicating that every model exhibited an additional loss across the 16 target languages even after its own English--Portuguese gap was subtracted.

Most of the raw gap nevertheless corresponded to the difference between English and Portuguese. This component ranged from 0.050 for GPT-5.6 Terra {[}0.043, 0.056{]} to 0.079 for Claude Sonnet 5 {[}0.070, 0.088{]}. The identity among the three components held for every model:

\[
\mathrm{Raw\ CLCG}
=
\mathrm{EN\text{–}PT\ Gap}
+
\mathrm{Net\ CLCG}.
\]

The complete model-level decomposition, including component-specific confidence intervals, is reported in Supplementary Figure S1 and Supplementary Material Section SM7.4. That section also documents the 0.001 display-level rounding residual for Gemini 3.6 Flash, which arises because the components and raw total are rounded independently.

\begin{figure}
\centering
\pandocbounded{\includegraphics[width=0.80\linewidth,keepaspectratio]{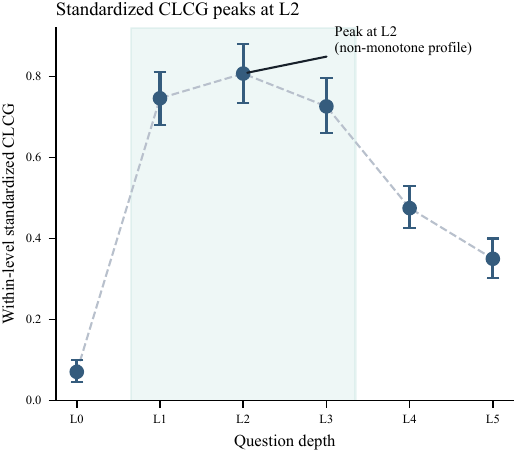}}
\caption{Within-level standardized CLCG by question depth. Points show pooled CLCG estimates after scores were standardized within each question level, and vertical bars show 95\% percentile confidence intervals from 2,000 bootstrap resamples clustered by article. The thin dashed connector is only a visual guide across the ordered levels and does not imply equal spacing or a continuous psychometric scale. The profile peaks at \texttt{L2} and is nonmonotonic. The positive average slope of 0.014 {[}0.004, 0.025{]} (\texttt{p\ =\ 0.002}) summarizes the overall ordered trend and does not imply monotonic growth between consecutive levels. The non-monotonic profile and its \texttt{L2} peak are reproduced under an independent hybrid re-labeling of the items (Supplementary Material Section SM11).}
\label{fig:standardized-depth-profile}
\end{figure}

Against the natural expectation that the gap should widen with cognitive depth, Figure~\ref{fig:standardized-depth-profile} shows that the level-specific profile did the opposite: it peaked in the middle and declined toward the deepest items. After standardizing scores within each level, the CLCG was 0.071 for \texttt{L0}, rose to 0.746 for \texttt{L1} and 0.807 for \texttt{L2}, then fell across 0.726 for \texttt{L3}, 0.475 for \texttt{L4}, and 0.350 for \texttt{L5}. The largest contrasts thus concentrated in the literal-extraction and reorganization band (\texttt{L1}--\texttt{L3}), not in the deepest integrative tasks.

Despite this nonmonotonic pattern, the overall estimated slope across depth levels was positive, with a mean of 0.014, a 95\% confidence interval of 0.004 to 0.025, and \texttt{p\ =\ 0.002}. This estimate represents an average trend in the standardized contrast rather than continuous growth between consecutive levels.

The raw level-specific values were 0.024 for \texttt{L0}, 0.242 for \texttt{L1}, 0.166 for \texttt{L2}, 0.126 for \texttt{L3}, 0.063 for \texttt{L4}, and 0.044 for \texttt{L5}. Because \texttt{L0} uses Accuracy whereas the open-ended levels use Token-F1, these values are presented descriptively and should not be compared directly.

Exact standardized estimates for \texttt{L0}--\texttt{L5} are reported in Supplementary Material Section SM7.4. The model-level decomposition expands the raw model comparison without duplicating the main-paper forest plot.

\subsubsection{Robustness Across Domains}\label{robustness-across-domains}

The CLCG was also observed outside the primary corpus. In FLORES-200, using \texttt{L2}--\texttt{L3} items and three of the evaluated models, the pooled CLCG was 0.136. When the analysis was restricted to \texttt{L3}, the pooled value increased to 0.156. These results show that the cross-lingual loss is not confined to the editorial domain or textual genre of the primary corpus.

\begin{figure}
\centering
\pandocbounded{\includegraphics[width=\linewidth,keepaspectratio]{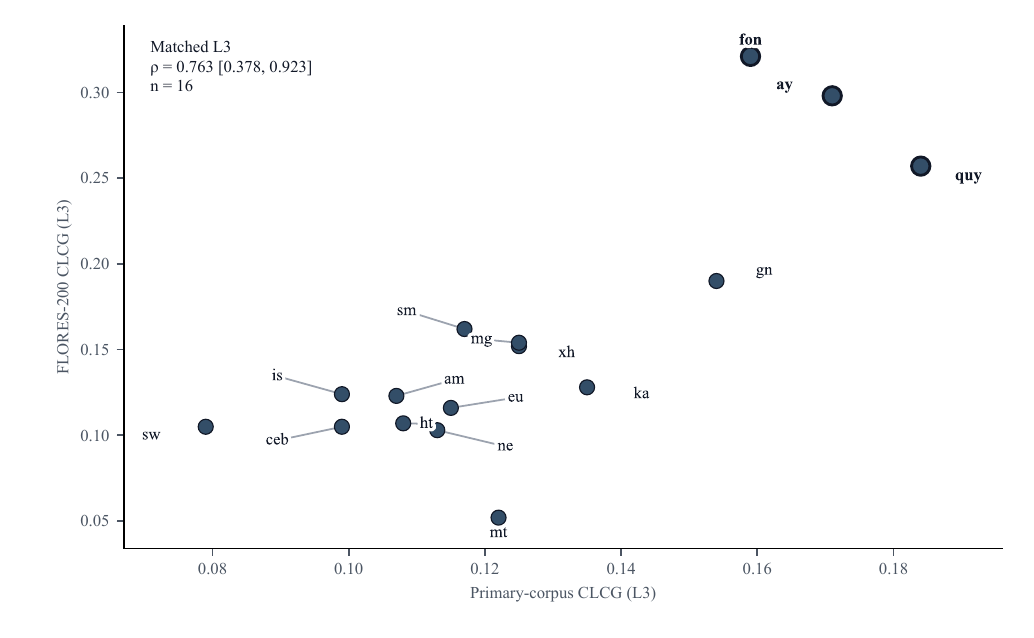}}
\caption{Cross-domain robustness of language-level CLCG. Points compare matched-\texttt{L3} language-level CLCG estimates in the primary corpus and FLORES-200 across the 16 target languages (Goyal et al., 2022; NLLB Team, 2024). The matched-depth Spearman correlation was \texttt{\ensuremath{\rho}=0.763} (95\% CI \texttt{{[}0.378,\ 0.923{]}}, \texttt{n=16}). Fon, Aymara, and Ayacucho Quechua, which exhibited the largest FLORES-200 \texttt{L3} gaps, are highlighted. In the broader protocol comparison, the primary-corpus headline \texttt{L3}--\texttt{L5} ranking correlated \texttt{\ensuremath{\rho}=0.791} (95\% CI \texttt{{[}0.428,\ 0.931{]}}) with the FLORES \texttt{L2}--\texttt{L3} ranking, and the FLORES split-half correlation was \texttt{0.926}. Only three models were evaluated in FLORES-200. Because the corpora differ in domain, text length, item construction, and model coverage, absolute gap magnitudes should not be interpreted as directly equivalent. Complete language- and model-level values are reported in Supplementary Tables SM6.1 and SM6.2.}
\label{fig:cross-domain-flores}
\end{figure}

The primary visual comparison matched question depth exactly by comparing \texttt{L3} language-level estimates in the two corpora. Across the 16 target languages, the matched-depth Spearman correlation was 0.763 (95\% CI 0.378--0.923), as Figure~\ref{fig:cross-domain-flores} displays.

A broader protocol comparison contrasted the primary-corpus headline ranking from \texttt{L3}--\texttt{L5} with the FLORES ranking from \texttt{L2}--\texttt{L3}. Although the depth bands are not identical, the association was similarly strong at 0.791, with a 95\% confidence interval of 0.428 to 0.931. This secondary result shows that the cross-domain agreement was not dependent on the strict \texttt{L3} restriction, while the matched-depth comparison remains the cleaner visual test.

The three models included in the external analysis exhibited similar pooled \texttt{L2}--\texttt{L3} gaps: 0.135 for DeepSeek V4 Pro, 0.131 for Gemini 3.6 Flash, and 0.143 for Kimi K3. The proximity of these estimates indicates that the cross-lingual effect persisted across models, although the external analysis covered only a subset of the five-model primary panel. Pairwise agreement among their language rankings ranged from 0.880 to 0.985.

The external results also showed high internal stability. The split-half Spearman correlation between language-level estimates was 0.926. The largest aggregate \texttt{L2}--\texttt{L3} gaps occurred for Fon, Aymara, and Ayacucho Quechua, at 0.300, 0.281, and 0.238, respectively. These languages were also among those with the largest losses in the primary corpus.

The difference in magnitude between the corpora should not be interpreted as a direct comparison of difficulty. FLORES-200 differs from the primary corpus in text distribution, document length, item construction, and model coverage. The central evidence is the preservation of the direction of the effect and the relative ordering of languages under matched depth, rather than equality of absolute estimates.

Together, these results provide evidence of cross-domain robustness. The CLCG remains positive in an independent corpus, reproduces across models, and largely preserves the language hierarchy observed in the primary study.

\subsubsection{Stability Across Inference Repetitions}\label{stability-across-inference-repetitions}

To assess whether the CLCG could be explained by stochastic generation variability, we repeated inference three times on a subsample of 40 articles, holding constant the models, items, contexts, prompts, and scoring criteria.

The aggregate CLCG remained stable across the three repetitions, with values of 0.073, 0.079, and 0.070. The mean was 0.074, with a standard deviation of 0.005 and a total range of 0.009. This range was smaller than the width of the 95\% confidence interval for the primary result, which was 0.012, indicating that decoding variability remained within the previously estimated sampling uncertainty.

Stability was also observed in the model rankings. Overall agreement among the rankings across the three repetitions, measured using Kendall's coefficient of concordance, was W = 0.911. The mean pairwise Spearman correlation between repetitions was 0.867, with individual values of 0.900, 0.800, and 0.900.

The magnitude of the variation was not identical across models. GPT-5.6 Terra showed the greatest stability, with a mean CLCG of 0.058, a standard deviation of 0.002, and a range of 0.004. Gemini 3.6 Flash and Kimi K3 also showed low dispersion, with ranges of 0.008 and 0.009, respectively. DeepSeek V4 Pro had a range of 0.019, whereas Claude Sonnet 5 showed the largest variation, with values of 0.100, 0.111, and 0.080, corresponding to a range of 0.031.

Even for Claude Sonnet 5, the CLCG remained positive in all repetitions. Generation variability therefore affected the point estimate of the effect for some models, but not its direction.

The rankings also showed minor position changes among models with similar values. GPT-5.6 Terra and Gemini 3.6 Flash exchanged positions across the first two repetitions, whereas DeepSeek V4 Pro and Kimi K3 exchanged positions in the third. These changes did not alter the central conclusion because all models retained positive CLCG values and overall ranking agreement remained high.

Together, these results show that the CLCG is not the product of a single decoding run. Both its aggregate magnitude and the general ordering of the models were reproduced across multiple executions, and the observed variation remained smaller than the sampling uncertainty of the primary estimator.

\begin{figure}
\centering
\pandocbounded{\includegraphics[width=0.80\linewidth,keepaspectratio]{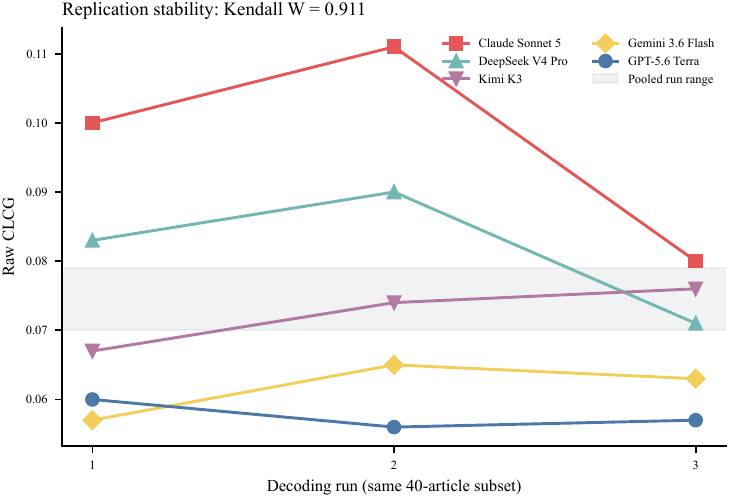}}
\caption{Stability of CLCG across repeated decoding runs. Lines show model-level CLCG in three decoding runs on the same 40-article subset; stored repetitions 0--2 are displayed as Runs 1--3. The shaded band is the observed pooled run range (0.070--0.079), not an inferential confidence interval. Kendall's coefficient of concordance across model rankings was \texttt{W\ =\ 0.911}.}
\label{fig:replication-stability}
\end{figure}

Figure~\ref{fig:replication-stability} presents the CLCG across repetitions at both the aggregate and model levels and highlights the relative stability of the rankings.

\subsubsection{Memorability and Possible Contamination}\label{memorability-and-possible-contamination}

A natural worry is the opposite of the one usually raised: rather than inflating the gap, prior exposure to the texts might quietly deflate it. To test this, we classified each model--article pair as potentially memorizable or new according to the relationship between the article's earliest availability date and the model's knowledge cutoff. The distribution of articles published before and after the cutoff varied substantially across models. The complete table of cutoffs and coverage is provided in Supplementary Material Section SM9.4. The result runs counter to the simplest contamination story: the gap was \emph{smaller} for content the models had most likely already seen, so evaluations built on older, familiar text may \emph{understate} the true cross-lingual difficulty rather than manufacture it.

In the aggregate analysis, the CLCG was 0.076 for potentially memorizable content, with a 95\% confidence interval of 0.071 to 0.083, and 0.086 for new content {[}0.074, 0.098{]}. The difference between the two conditions was \ensuremath{-}0.009 {[}\ensuremath{-}0.020, 0.000{]}, with a one-sided test yielding \texttt{p\ =\ 0.028}.

The negative sign indicates that the CLCG was smaller for potentially memorizable content. In other words, prior familiarity with the text may partially reduce the observed loss between English and the target languages. This pattern is consistent with the hypothesis that evaluations based exclusively on content widely available before the cutoff may underestimate the actual difficulty of cross-lingual comprehension.

The mixed-effects model produced a convergent result. The interaction between the English condition and memorability was \ensuremath{-}0.009, with a 95\% confidence interval of \ensuremath{-}0.017 to \ensuremath{-}0.002 and \texttt{p\ =\ 0.015}. The article-level block bootstrap and the crossed-effects model therefore pointed in the same direction.

The model-specific analysis showed heterogeneity. For GPT-5.6 Terra, the CLCG was 0.060 for potentially memorizable content and 0.079 for new content, yielding a difference of \ensuremath{-}0.019 {[}\ensuremath{-}0.033, \ensuremath{-}0.004{]} and \texttt{p\ =\ 0.007}. This was the clearest model-specific contrast.

For Gemini 3.6 Flash, the difference was larger in magnitude at \ensuremath{-}0.024, but the confidence interval included zero {[}\ensuremath{-}0.061, 0.004{]}, with \texttt{p\ =\ 0.051}. The difference was \ensuremath{-}0.008 {[}\ensuremath{-}0.031, 0.011{]} for Claude Sonnet 5 and \ensuremath{-}0.009 {[}\ensuremath{-}0.026, 0.007{]} for Kimi K3. A within-model contrast could not be estimated for DeepSeek V4 Pro because all 150 articles in the sample were published before its knowledge cutoff.

The class-level sensitivity analysis showed negative memorizable-minus-novel contrasts in every estimable Joshi class. The estimates were \ensuremath{-}0.026 {[}\ensuremath{-}0.041, \ensuremath{-}0.013{]} in Class 0, \ensuremath{-}0.011 {[}\ensuremath{-}0.022, \ensuremath{-}0.000{]} in Class 1, \ensuremath{-}0.006 {[}\ensuremath{-}0.016, 0.004{]} in Class 2, \ensuremath{-}0.014 {[}\ensuremath{-}0.026, \ensuremath{-}0.003{]} in Class 3, and \ensuremath{-}0.004 {[}\ensuremath{-}0.016, 0.008{]} in Class 4. Because Class 0 contains only Fon and several classes include few languages, these estimates are descriptive rather than confirmatory.

These results do not demonstrate direct memorization of specific answers. The classification uses only a temporal condition indicating the possibility of exposure. An article published before the cutoff is considered potentially memorizable, but this does not establish that the model encountered or retained that content during training. The cutoffs also vary in precision, and the cutoff for Kimi K3 is explicitly approximate.

The appropriate interpretation is therefore a sensitivity analysis concerning possible contamination. The CLCG remained positive for both potentially memorizable and new content but was modestly larger in the new-content subset. Prior familiarity may therefore attenuate the magnitude of the effect without fully explaining it.

Together, the analysis suggests that the primary result is not produced by training-data contamination. Instead, when the study is restricted to content published after the cutoff of models for which such a comparison is possible, the CLCG tends to increase.

\begin{figure}
\centering
\pandocbounded{\includegraphics[width=\linewidth,keepaspectratio]{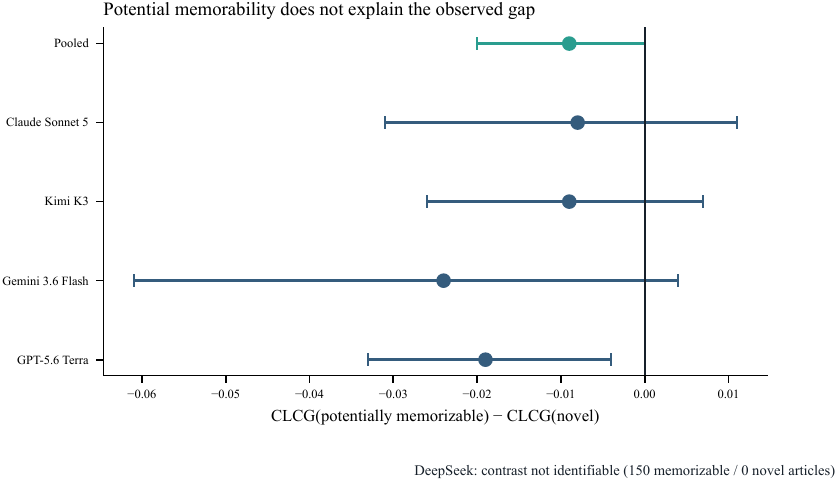}}
\caption{Sensitivity of CLCG to potential memorability. Negative values indicate a smaller CLCG for potentially memorizable articles than for articles first available after the model cutoff. Bars are stored 95\% confidence intervals. DeepSeek V4 Pro is not plotted because its 150/0 article split makes the within-model contrast non-identifiable. The temporal classification indicates possible exposure, not verified training-data membership or memorization.}
\label{fig:memorability-sensitivity}
\end{figure}

Figure~\ref{fig:memorability-sensitivity} compares the potentially-memorizable-minus-new CLCG contrast in the aggregate and by model. Class-level estimates remain available in the SoT and Supplementary Material but are not duplicated in the figure.

\subsubsection{Context Gain and the Distinction Between Reading and Prior Knowledge}\label{context-gain-and-the-distinction-between-reading-and-prior-knowledge}

To investigate whether the CLCG reflected difficulty in reading the provided context or only differences in the models' prior knowledge, we compared performance with the passage available against performance in the \texttt{closed\_book} condition, in which the question was presented without its corresponding passage.

Mean performance in the \texttt{closed\_book} condition was 0.285. Because this condition contained no passage in any language, it served as a common baseline for prior knowledge and the ability to answer without consulting the text.

We defined context gain as the difference between the score obtained with the passage available and the score obtained in the \texttt{closed\_book} condition. The mean gain was 0.147 in English and 0.087 in Portuguese. Under the Joshi classification, mean gain was 0.006 in Class 0, 0.053 in Class 1, 0.076 in Classes 2 and 3, and 0.089 in Class 4. Class 0 is represented only by Fon, and Ayacucho Quechua is excluded from primary class aggregates.

This pattern indicates that the models extracted more useful information from the English context than from contexts in lower-resource languages. The difference cannot be attributed solely to a general ability to answer the question because the \texttt{closed\_book} baseline was common across languages. What varied was the additional benefit provided by reading the passage.

The language-level analysis reinforced this interpretation. The largest context gains occurred in English (0.147), Basque (0.091), Swahili (0.090), Portuguese (0.087), and Samoan (0.087). In contrast, the smallest gains were observed in Fon (0.006), Ayacucho Quechua (0.016), and Aymara (0.026).

Fon, Aymara, and Ayacucho Quechua were also among the languages with the largest CLCG values. The combination of a large cross-lingual gap and a small context gain suggests that the models derived little additional benefit from receiving the passage in these languages.

The \texttt{closed\_book} condition does not, however, demonstrate direct memorization. A correct answer without the text may result from prior factual knowledge, plausible inference, familiarity with the topic, or previous exposure to the content. This analysis is therefore interpreted as a decomposition between performance without context and the benefit of reading, rather than as a definitive test of contamination.

The \texttt{closed\_book} analysis was exploratory and used only 36 items, corresponding to 143 valid responses from levels \texttt{L3}--\texttt{L5}. The frozen score file contains four of the five evaluated models and does not include \texttt{claude-sonnet-5}; the analysis is therefore a partial-panel diagnostic rather than a fully balanced five-model comparison. Its purpose is to provide complementary mechanistic evidence rather than to replace the primary estimator based on the full confirmatory analytical sample. The implemented inner join yields 2,455 matched model--language--item rows: 24 for Basque and 143 for every other language. The source of truth stores no confidence interval, standard error, hypothesis test, multiplicity adjustment, or context-gain breakdown by model, depth, or memorability. The complete frozen values and coverage audit are reported in Supplementary Material Section SM4.

Despite this limitation, the direction of the result is consistent with the central hypothesis. If the CLCG were driven primarily by differences in prior knowledge, the condition without context would be expected to explain much of the variation across languages. Instead, the benefit of the passage decreased precisely in the language groups with the largest gaps.

Together, these results indicate that the CLCG is associated with an unequal ability to extract information from semantically equivalent passages across languages. The models not only perform worse in lower-resource languages, but also derive less benefit when explicitly provided with the text required to answer the question.

\begin{figure}
\centering
\pandocbounded{\includegraphics[width=\linewidth,keepaspectratio]{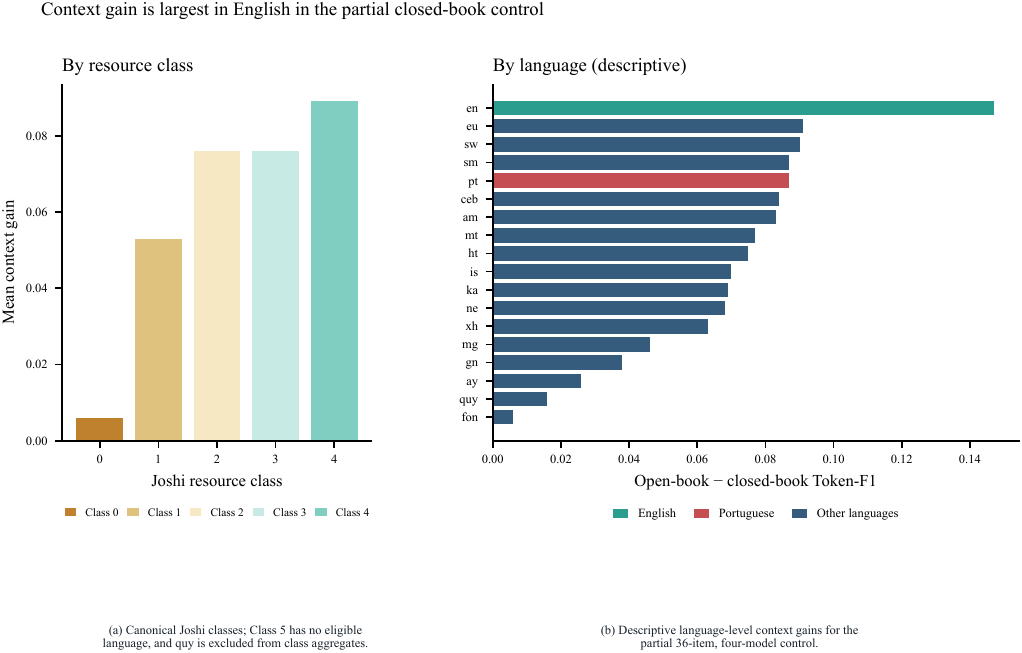}}
\caption{Closed-book context gain by language and resource class. Bars show descriptive open-book-minus-closed-book Token-F1 estimates for the partial 36-item, four-model control (143 valid responses). The SoT stores no inferential interval for this diagnostic, so none is shown. Ayacucho Quechua is retained in the language panel but excluded from canonical resource-class aggregation.}
\label{fig:context-gain-profile}
\end{figure}

Figure~\ref{fig:context-gain-profile} presents context gain by language and resource class, highlighting the reduction observed among lower-resource languages.

\subsubsection{Factors Associated With Variation Across Languages}\label{factors-associated-with-variation-across-languages}

The magnitude of the CLCG varied substantially across the 16 target languages. The executed factor analysis examined canonical Joshi resource class, \texttt{cl100k\_base} tokenizer fertility, a binary non-Latin-script indicator, and a binary isolate or quasi-isolate probe. Portuguese and the 15 targets with canonical classes formed the primary language-level sample; Ayacucho Quechua entered only the explicit proxy sensitivity.

\begin{figure}
\centering
\pandocbounded{\includegraphics[width=\linewidth,keepaspectratio]{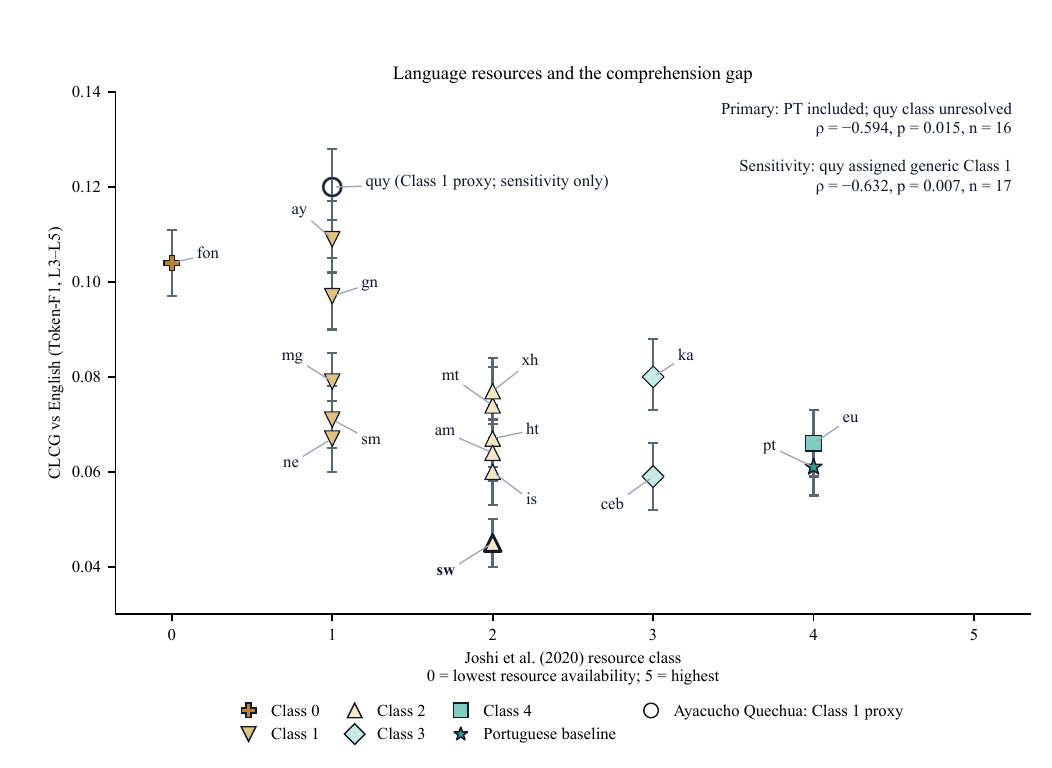}}
\caption{Language resources and the comprehension gap. Points show language-level CLCG estimates in the primary \texttt{L3}--\texttt{L5} Token-F1 band, with vertical bars representing 95\% confidence intervals obtained through bootstrap resampling clustered by article. The primary Spearman correlation includes Portuguese and the 15 target languages with an unambiguous Joshi et al.~(2020) classification (\texttt{\ensuremath{\rho}\ =\ -0.594}, 95\% CI {[}-0.848, -0.138{]}, \texttt{p\ =\ 0.015}, \texttt{n\ =\ 16}). Ayacucho Quechua is retained in the figure but has no canonical class in the primary mapping; its open marker at the generic Class 1 assignment represents only the sensitivity analysis (\texttt{\ensuremath{\rho}\ =\ -0.632}, 95\% CI {[}-0.838, -0.228{]}, \texttt{p\ =\ 0.007}, \texttt{n\ =\ 17}). Higher class values indicate greater resource availability, so the negative coefficients denote larger gaps among lower-resource languages. Class 5 is shown on the axis but contains no analyzed observation. The association is descriptive and does not imply that resource availability causally determines the CLCG.}
\label{fig:resource-rank-vs-clcg}
\end{figure}

The largest bivariate association among the executed probes involved the externally defined Joshi resource class. Ayacucho Quechua was retained in all language-level results but left unclassified in the primary Joshi mapping because its specific variety could not be resolved unambiguously. The primary Spearman correlation therefore used Portuguese and the 15 target languages with canonical classes and was -0.594 (\texttt{p\ =\ 0.015}, \texttt{n\ =\ 16}; 95\% CI {[}-0.848, -0.138{]}). The negative sign is expected because higher class numbers denote greater resource availability, whereas larger CLCG values denote greater loss. Figure~\ref{fig:resource-rank-vs-clcg} displays all language-level observations, including Ayacucho Quechua as an explicitly labeled sensitivity proxy.

The class means were not perfectly monotonic because the classes are ordinal, small, and unevenly populated. Mean CLCG was 0.104 in Class 0, 0.085 in Class 1, 0.065 in Class 2, 0.070 in Class 3, and approximately 0.063 in Class 4. Class 0 contained only Fon, and no analyzed target language belonged to Class 5.

A sensitivity analysis assigning the generic Quechua mapping to Ayacucho Quechua as Class 1 produced a similar association (\texttt{\ensuremath{\rho}=-0.632}, \texttt{p=0.007}, \texttt{n=17}; 95\% CI {[}-0.838, -0.228{]}). This proxy assignment was used only for the sensitivity analysis; it did not reclassify Ayacucho Quechua in the primary mapping. The small change in magnitude did not alter the direction or interpretation of the result.

Other factors showed smaller and more uncertain associations. Tokenizer fertility had a positive correlation of 0.455 (\texttt{p\ =\ 0.077}; language-bootstrap 95\% CI {[}0.006, 0.742{]}) but did not meet the executed approximate guide of \texttt{\textbar{}\ensuremath{\rho}\textbar{}\ \ensuremath{\geq}\ 0.48}; the available provenance does not establish that guide as prespecified. The correlation was 0.000 (\texttt{p\ =\ 1.000}; 95\% CI {[}-0.414, 0.429{]}) for the non-Latin indicator and 0.082 (\texttt{p\ =\ 0.763}; 95\% CI {[}-0.332, 0.495{]}) for the isolate or quasi-isolate probe.

After rank residualization on canonical Joshi class, the partial association was 0.100 (\texttt{p\ =\ 0.713}, \texttt{n\ =\ 16}) for fertility and 0.122 (\texttt{p\ =\ 0.653}, \texttt{n\ =\ 16}) for non-Latin script. These partial analyses have no stored confidence intervals. The standardized exploratory OLS had \texttt{R\textsuperscript{2}\ =\ 0.460}; its coefficients were \texttt{-0.463} for Joshi class, \texttt{0.939} for fertility, and \texttt{-0.946} for non-Latin script. Because the model stores no inferential standard errors or intervals and fertility correlates \texttt{\ensuremath{\rho}\ =\ 0.677} with the non-Latin indicator, these coefficients are descriptive and are not interpreted as isolated effects.

The executed probes do not support reducing CLCG variation to either the binary writing-system indicator or the isolate-like indicator. Among the measured variables, Joshi class had the largest bivariate association. It is nevertheless an aggregate proxy that may reflect multiple correlated aspects of digital representation and historical investment in language technologies.

The association with resource availability should not be interpreted causally. The available measures are aggregate proxies and are correlated with one another. Lower-resource languages may also differ in tokenization, geographic representation, digital availability, diversity of textual genres, and the quality of the data used to train the models.

Variation among languages within the same class is also informative. Swahili, for example, exhibited a relatively small CLCG within Class 2, whereas Xhosa showed a larger gap in the same class. Within Class 1, Aymara and Guarani had larger gaps than Malagasy, Nepali, and Samoan. These differences show that resource classification explains part, but not all, of the observed heterogeneity.

Within this small language panel, the results show a descriptive inverse association between Joshi resource class and cross-lingual performance loss. The pattern is probabilistic rather than deterministic and does not establish that resource class independently determines the behavior of a language.

The full language inventory, bivariate and partial associations, predictor-correlation matrix, descriptive OLS, and non-executed-analysis boundary are reported in Supplementary Material Sections SM6.12--SM6.14. Figure~\ref{fig:bivariate-language-factors} summarizes the exploratory bivariate associations among the executed language-level probes.

\begin{figure}
\centering
\pandocbounded{\includegraphics[width=\linewidth,keepaspectratio]{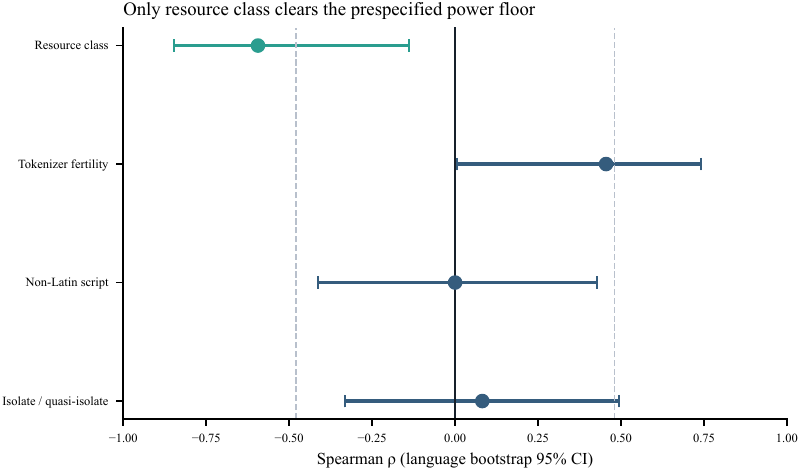}}
\caption{Exploratory bivariate language-factor associations. Points show current bivariate Spearman coefficients and language-bootstrap 95\% confidence intervals for Portuguese plus 15 canonically classified targets (\texttt{n\ =\ 16}). Dashed lines mark the executed approximate guide \texttt{\textbar{}\ensuremath{\rho}\textbar{}\ =\ 0.48}; only Joshi resource class exceeded that magnitude. The plot is descriptive, omits the non-inferential OLS, and does not duplicate the resource-class scatterplot.}
\label{fig:bivariate-language-factors}
\end{figure}

\subsubsection{Human Triangulation of the CLCG and Automatic Metrics}\label{human-triangulation-of-the-clcg-and-automatic-metrics}

Human data collection concluded after 200 participants had been recruited. Following application of the quality criteria, 21 submissions were excluded because of failed attention checks and two because of an exceptionally high number of internal contradictions across rubric fields. An additional 42 evaluations were removed only at the item level. The final analytical sample retained 177 annotators, 665 items, 10,407 dimension-level records, 138 decisive paired judgments, and 775 observations that could be directly compared with both automatic error judges.

The automatic metrics showed positive but modest associations with human judgments. Token-F1 correlated with adequacy at \texttt{\ensuremath{\rho}=0.232} (\texttt{p\textless{}0.001}, \texttt{n=564}) and with \texttt{keypoint\_recall} at \texttt{\ensuremath{\rho}=0.285} (\texttt{p\textless{}0.001}, \texttt{n=564}). The stronger correlation with factual coverage is consistent with Token-F1 being sensitive to the presence or absence of lexical content. However, the magnitudes indicate that neither human measure is well approximated by a single automatic metric.

Class-level human results were directionally compatible with the automatic pattern but highly imprecise. Across the five non-English Joshi classes represented in the human sample, the correlation between mean human adequacy and automatic CLCG was \texttt{\ensuremath{\rho}=-0.600} (\texttt{p=0.285}, \texttt{n=5}). The negative sign is expected because classes with larger automatic gaps tend to receive lower human adequacy ratings. The small number and unequal composition of classes, including a singleton Class 0 and no target language in Class 5, preclude strong inference.

The paired arm provided the human evidence most directly aligned with the study's primary contrast. Across 138 decisive judgments, the higher-resource response was preferred in 85 cases, corresponding to a raw proportion of 0.616 (exact 95\% CI {[}0.529, 0.697{]}; one-sided exact \texttt{p=0.004}). Because judgments were repeated within annotators and response pairs, the primary inference used a crossed-random-intercept logistic model. The estimated probability of preferring the higher-resource response was 0.655 (95\% CI {[}0.558, 0.741{]}; one-sided \texttt{p=0.001}; odds ratio 1.902, 95\% CI {[}1.262, 2.866{]}). An annotator-cluster bootstrap produced a similar estimate of 0.616 (95\% CI {[}0.535, 0.696{]}).

The adequacy scale showed upper-end concentration but did not trigger the executed complete-saturation flag. Median ratings were 4 in every Joshi class, below the rule's 4.5 median threshold. Responses in the HIGH automatic-score range received a mean adequacy rating of 4.216, compared with 3.819 in the LOW range, a positive difference of 0.397 that remained below the executed 0.5 separation cutoff. Thus, the flag remained false because the median condition failed; the result does not establish strong HIGH--LOW discrimination under the executed cutoff.

The principal limitation was low inter-rater agreement. Krippendorff's \ensuremath{\alpha} was 0.072 for adequacy and 0.154 for \texttt{keypoint\_recall}. All other dimensions were also far below the prespecified threshold of 0.67, ranging from \texttt{\ensuremath{-}0.034} to \texttt{0.124}. Aggregate human means may therefore reveal population-level tendencies, but individual labels and small differences should not be treated as stable human reference judgments.

Correspondence between human judgments and the automatic error judges was also low. Exact agreement with DeepSeek was 47.5\%, with \texttt{\ensuremath{\kappa}=0.131}, and exact agreement with Mistral was 48.4\%, with \texttt{\ensuremath{\kappa}=0.127}. When the two automatic judges agreed with each other, the human judgment confirmed their consensus in 55.3\% of cases. These results indicate that the automatic taxonomy is useful as an aggregate descriptive diagnostic but should not be presented as a human-validated classification.

The direction of the gap remained stable when the analysis was restricted to items included in the human study. Human and automatic scores were also positively correlated across all three item sources: \texttt{\ensuremath{\rho}=0.166} for pre-existing human-authored questions, \texttt{\ensuremath{\rho}=0.228} for shallow LLM-generated items, and \texttt{\ensuremath{\rho}=0.284} for deep LLM-generated items. Together, these results provide partial triangulation. Human evaluators perceive an advantage under higher-resource conditions, and their judgments modestly track the automatic scores. However, the low inter-rater agreement prevents the study from being interpreted as strong validation or definitive calibration of the CLCG.

Among 1,543 post-QC single-item judgments eligible for both measures, 224 (14.5\%) combined the maximum adequacy rating with incomplete atomic-fact coverage, affecting 193 of 598 eligible items (32.3\%). These judgments were retained in the primary analysis because global semantic adequacy and exhaustive factual coverage are related but non-identical constructs. Post-hoc removal of either the 224 affected annotator--item judgments or all 193 affected items changed several descriptive magnitudes: adequacy decreased and factual-coverage estimates increased. The Token-F1 associations remained positive and statistically supported, Krippendorff's \ensuremath{\alpha} remained far below 0.67, and the paired arm was unchanged. The sensitivities therefore demonstrate selective sample change rather than numerical invariance.

\begin{figure}
\centering
\pandocbounded{\includegraphics[width=\linewidth,keepaspectratio]{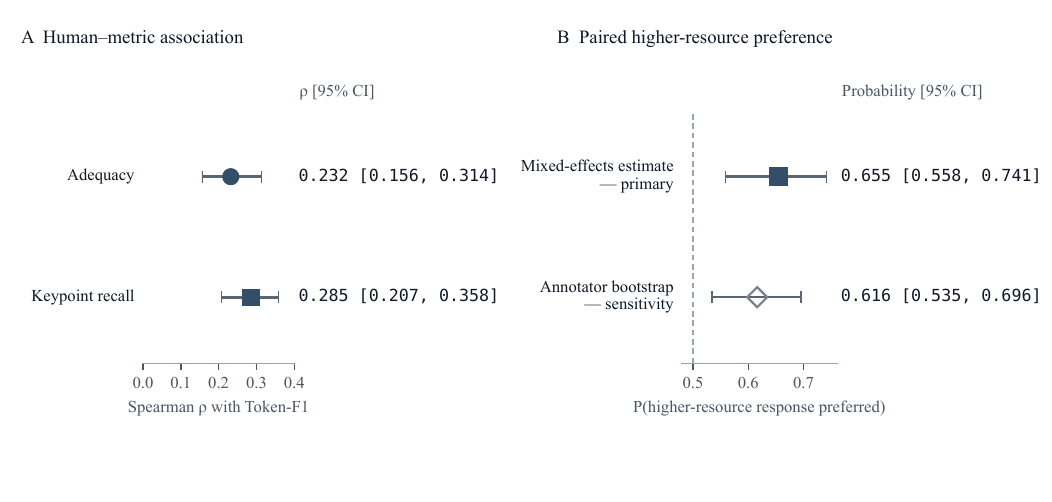}}
\caption{Human triangulation of the CLCG. Panel A shows Spearman associations between Token-F1 and human adequacy and keypoint recall. Horizontal bars are 95\% percentile confidence intervals from 2,000 bootstrap resamples clustered by item; both associations have \texttt{p\textless{}0.001} and \texttt{n=564}. Panel B shows the estimated probability that the response produced under the higher-resource language condition was preferred. The filled square is the primary crossed-random-intercept logistic-model estimate, and the open diamond is the annotator-cluster bootstrap sensitivity estimate; the dashed line marks chance probability \texttt{0.50}. The raw result was \texttt{85/138\ =\ 0.616} and is reported with the complete inferential details in Supplementary Table SM5.3. The positive but modest associations and the paired preference support partial triangulation, not calibration of the automatic metrics against a stable human gold standard. The quality-control flow is reported in Supplementary Table SM5.1, and the low agreement coefficients are shown in Supplementary Figure S2 and Supplementary Table SM5.4.}
\label{fig:human-triangulation}
\end{figure}

Figure~\ref{fig:human-triangulation} summarizes the two principal forms of positive human evidence: modest human--metric association and paired preference for the higher-resource condition.

\subsubsection{Error Patterns Associated With Cross-Lingual Loss}\label{error-patterns-associated-with-cross-lingual-loss}

To characterize the types of failure associated with the CLCG, we analyzed a stratified sample of 2,000 English--target-language response pairs. The sampling procedure deliberately increased the representation of cases with possible cross-lingual loss and limited the concentration of items from any single article. The frequencies reported in this section therefore describe the diagnostic sample and should not be interpreted as prevalence estimates for the full scored dataset.

Each side of the pair was classified by two judges from distinct model families, DeepSeek V4 Flash and Mistral Small. The analysis was kept separate from the primary CLCG estimator and was used only to identify patterns associated with score differences.

According to the DeepSeek judge, 64.4\% of target-language responses were classified as correct. The most frequent error was \texttt{partial\_or\_omission}, observed in 22.9\% of cases, followed by \texttt{unsupported\_addition} at 7.0\%. Contradictions accounted for 1.5\%, format or parsing failures for 1.4\%, and each remaining category occurred in fewer than 1\% of the pairs.

The most common transition was \texttt{correct\ \ensuremath{\rightarrow}\ correct}, with 1,134 pairs, indicating that a large proportion of responses remained correct under both language conditions. Among transitions representing deterioration, the most frequent was \texttt{correct\ \ensuremath{\rightarrow}\ partial\_or\_omission}, with 246 cases, followed by \texttt{correct\ \ensuremath{\rightarrow}\ unsupported\_addition}, with 102, \texttt{correct\ \ensuremath{\rightarrow}\ contradiction}, with 25, and \texttt{correct\ \ensuremath{\rightarrow}\ format\_or\_parse\_failure}, with 24.

These results indicate that cross-lingual loss occurred primarily as partial degradation rather than as a widespread shift to entirely incorrect responses. In many cases, the model preserved some correct information but omitted elements required to cover the gold answer fully. Unsupported additions formed a second relevant pattern, suggesting that difficulty extracting evidence from the context may also increase reliance on inferences not supported by the passage.

The contribution of an error category to the total score difference depends on both its frequency and the mean magnitude of the associated loss. Responses classified as correct in the target language still made a positive contribution to the aggregate delta, reflecting gradual differences in completeness or formulation that were not captured by the discrete category. Among the error categories, \texttt{partial\_or\_omission} contributed approximately 0.047 to the cumulative delta in the sample, whereas \texttt{unsupported\_addition} contributed 0.024, contradictions 0.010, and format failures 0.009.

Error patterns also varied across Joshi resource classes. The cross-lingual loss rate estimated by the primary judge was 33.9\% in Class 0, 20.4\% in Class 1, 14.8\% in Class 2, 20.0\% in Class 3, and 19.0\% in Class 4. The progression was not monotonic, and Class 0 contains only Fon; the class comparison is therefore descriptive. In Class 0, 47.9\% of responses were classified as correct, while partial omission accounted for 26.4\%, unsupported addition for 11.6\%, and contradiction for 5.0\%.

The loss rate also varied across models, ranging from 16.5\% for Gemini 3.6 Flash to 23.8\% for GPT-5.6 Terra. These values did not reproduce the exact model ordering observed for the continuous CLCG. This divergence is expected because the error sample was stratified, the classifier uses discrete categories, and the primary CLCG is calculated from continuous score differences across the full primary analytical sample. These rates should therefore not be interpreted as an alternative model ranking.

The two judges estimated similar aggregate cross-lingual loss rates: 19.8\% for DeepSeek and 20.5\% for Mistral. Their respective 95\% confidence intervals were {[}0.173, 0.225{]} and {[}0.180, 0.235{]}. Binary agreement on whether cross-lingual loss occurred was 81.3\%, with \texttt{\ensuremath{\kappa}\ =\ 0.419}.

Exact agreement on the detailed categories was 76.4\%, with \texttt{\ensuremath{\kappa}\ =\ 0.462}. Although the observed agreement appears high, kappa remained below 0.50, and several rare categories showed little or no category-specific agreement. We therefore did not consolidate the judges' fine-grained labels into a definitive taxonomy.

The automatic error analysis should be interpreted as associative and descriptive evidence. Both judges are language models, one belongs to the same family as an evaluated model and the item generator, and the categorical classifications do not provide a causal explanation for the performance differences.

Nevertheless, the convergence between the two judges on the aggregate loss rate and the predominance of partial omissions indicate that the CLCG does not manifest only through format or parsing failures. Most of the observed deterioration was semantic, involving loss of relevant information, incomplete coverage of the gold answer, and, to a lesser extent, the introduction of unsupported content.

\begin{figure}
\centering
\pandocbounded{\includegraphics[width=\linewidth,keepaspectratio]{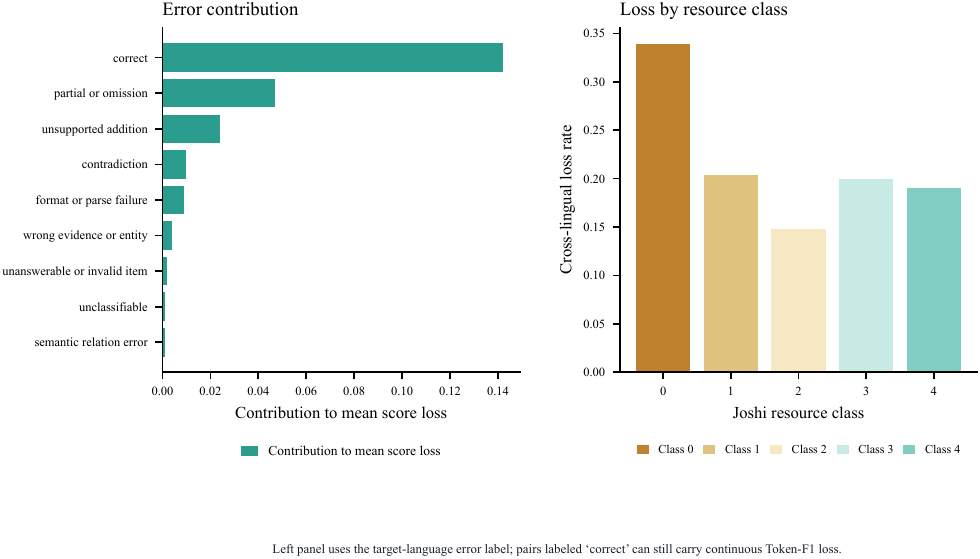}}
\caption{Cross-lingual error patterns and contribution. The left panel combines label frequency and mean continuous score difference in the 2,000-pair stratified sample. A target-language response labeled \texttt{correct} can still carry a Token-F1 difference, so the discrete label and continuous loss are not interchangeable. The right panel is descriptive; Class 0 contains only Fon and the class pattern is nonmonotonic.}
\label{fig:crosslingual-error-patterns}
\end{figure}

Figure~\ref{fig:crosslingual-error-patterns} presents each target-language error label's contribution to continuous score loss and the cross-lingual loss rate by Joshi resource class.

\subsubsection{Robustness Across Metrics and Analytical Specifications}\label{robustness-across-metrics-and-analytical-specifications}

The primary result was defined using micro-averaged Token-F1, while holding the article, question, gold answer, model, prompt, and context constant between the English and target-language conditions. To assess whether the CLCG depended on this specific choice, we repeated the analyses using alternative metrics and aggregation procedures.

The direction of the effect remained stable across the evaluated metrics. In addition to Token-F1, we considered Exact Match, accuracy derived from coverage of the gold-answer elements, semantic similarity, BERTScore, and entailment-based evaluation. The magnitudes were not directly comparable because each metric operates on a different scale and responds differently to paraphrases, omissions, and additional content. Nevertheless, languages exhibiting larger losses according to Token-F1 also tended to exhibit larger losses under the semantically oriented metrics.

Exact Match produced smaller absolute values and a greater concentration of zeros, as expected for open-ended responses. Because it requires much stricter textual correspondence, this metric was less sensitive to semantically correct responses expressed using different wording. It was therefore treated as a robustness analysis rather than as the primary estimator.

The semantic metrics reduced some of the penalties caused by lexical reformulation but did not eliminate the cross-lingual gradient. This result indicates that the CLCG does not arise solely from superficial differences between the model's wording and the reference answer. Even when the evaluation permits greater semantic flexibility, a systematic loss remains in lower-resource languages.

We also compared micro- and macro-aggregation. Micro-averaging assigns weight to each evaluation unit and constitutes the study's primary estimator. Macro-averaging assigns equal weight to groups such as languages, articles, or models, depending on the specification. Although the aggregate magnitude varied, the overall direction of the effect and the identification of the languages with the largest gaps remained consistent.

Analyses stratified by model, comprehension depth, Joshi resource class, and question source also preserved the predominant direction of the CLCG. Heterogeneity across strata affected the magnitude of the gap but provided no evidence that the primary result was restricted to a single model, difficulty level, or subset of items.

The use of Portuguese as an empirical baseline was also subjected to sensitivity analysis. The raw gap relative to English was presented alongside the net gap calculated after subtracting the English--Portuguese difference. This decomposition reduced the magnitude attributed specifically to the target languages but did not eliminate the effect. For all five models, the confidence intervals for the net CLCG remained above zero.

Equivalence between English and Portuguese was not assumed. The executed equivalence test did not support the conclusion that the difference between these two conditions fell within the stored margin. Portuguese was therefore not interpreted as a neutral condition or as equivalent to English, but as an empirical reference with high resource availability.

This distinction is central to the interpretation of the net CLCG. The estimator does not claim that the entire English--Portuguese difference represents a general language-independent bias. It uses a higher-resource condition only to estimate how much of the loss observed in the target languages exceeds a difference already present between two widely represented languages.

The confirmatory mixed-effects models provided an additional check where they were identifiable and converged. The specifications included effects associated with articles and models, preserving the crossed structure of the observations and reducing the risk that the results were dominated by a small set of texts or systems. The converged models were interpreted alongside the descriptive and bootstrap analyses. The H2 memorability model did not converge and was therefore excluded as independent confirmatory support.

The article-level block bootstrap also protected the inference against dependence among questions extracted from the same document. The confidence intervals therefore did not treat questions originating from the same passage or article as completely independent evidence.

Together, these analyses indicate that the CLCG is not a consequence of a single metric, aggregation procedure, model family, or baseline choice. Its magnitude varies across specifications, but the central conclusion remains unchanged: the same content produces lower performance when presented in multiple lower-resource languages than when presented in English.

The complete chain of validity and robustness evidence is synthesized in Figure~\ref{fig:evidence-synthesis} below. Detailed numerical results and analytical boundaries are reported in Supplementary Material Sections SM4--SM8, SM10, and SM11 at the corresponding points in the analysis.

\begin{figure}
\centering
\pandocbounded{\includegraphics[width=\linewidth,keepaspectratio]{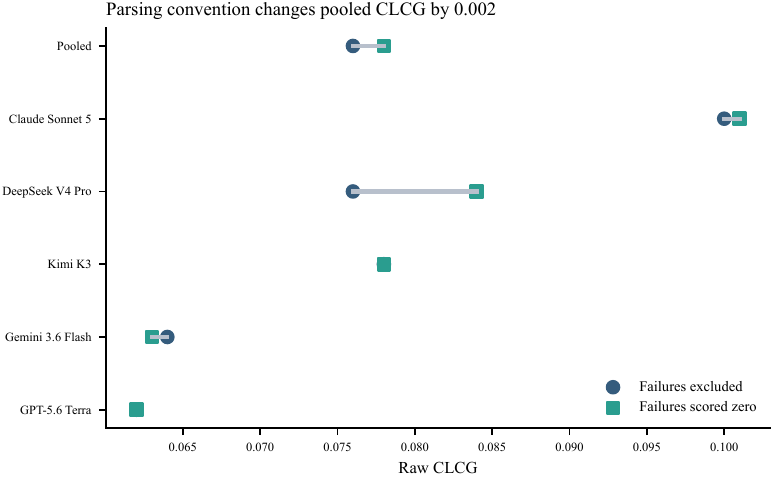}}
\caption{Robustness of CLCG to the parsing-failure convention. Paired points compare the two implemented rules in the primary \texttt{L3}--\texttt{L5} band. The 0.002 pooled shift is a specification contrast, not a confidence interval. The figure does not imply robustness to unexecuted alternative metrics.}
\label{fig:parsing-rule-robustness}
\end{figure}

Figure~\ref{fig:parsing-rule-robustness} summarizes the executed parsing-rule sensitivity, which provides a narrower specification check. Scoring failures as zero yielded pooled CLCG 0.078, while excluding 1,499 failed or missing parses yielded 0.076; all five model-level gaps and the pooled H1 direction remained positive.

A context-scope probe pointed in the same direction. Supplying only the linked paragraphs rather than the full article did not shrink the gap but enlarged it: on a matched 40-article population the CLCG rose from 0.118 under \texttt{full\_body} to 0.146 under \texttt{item\_grounded} ($\Delta=-0.028$, 95\% CI [$-0.037$, $-0.019$]), an effect present across all resource classes and larger for local than for multi-paragraph items. The gap is therefore not an artifact of long-document retrieval; if anything, the full article partially masks it. Full estimator, per-language and per-class values, and boundaries are in Supplementary Material Section SM10.

\subsection{Discussion}\label{discussion}

\subsubsection{The Models' Comprehension Is Not Linguistically Invariant}\label{the-models-comprehension-is-not-linguistically-invariant}

This study investigated whether language models preserve the same ability to answer questions about content when the information remains semantically comparable but the language of presentation changes. The results show that such invariance cannot be assumed.

The four confirmatory hypotheses can be summarized briefly. \textbf{H1} is supported: the pooled gap was positive (0.078, 95\% CI [0.072, 0.084]) and every one of the five models showed an individually positive gap whose interval excluded zero. \textbf{H2} is also supported, but in a way that cuts against a memorization account---the gap was \emph{smaller} for potentially memorizable content, which argues against prior exposure as its source. \textbf{H3} holds in the tested ordered-slope sense, with one important caveat: the standardized slope was positive, yet the raw profile was non-monotonic and peaked at \texttt{L2} rather than rising to the deepest levels. \textbf{H4} found that the gap does depend on how items were written---the LLM-generated and human-authored gaps were not equivalent---so the two sources are reported separately rather than pooled. The two results that run against the intuitive story, the non-monotonic depth profile and the reversed memorability effect, shape how the gap should be read.

The primary pooled scored-row CLCG relative to English was 0.078, while the equal-language macro summary across the 16 targets was 0.077. After subtracting the empirical difference observed between English and Portuguese at the language-summary level, a macro net gap of approximately 0.016 remained, with a confidence interval entirely above zero. The language-level gap was inversely associated with Joshi resource class (\texttt{\ensuremath{\rho}=-0.594}, \texttt{p=0.015}, \texttt{n=16}), although the class means were not monotonic and Class 0 was represented only by Fon. Two findings cut against the intuitive story and matter for how the gap should be read: it did not grow with cognitive depth but peaked in the middle of the range, and it was smaller---not larger---for content the models had likely encountered before, which points away from memorization as its source rather than toward it.

The absolute magnitude of the effect should be interpreted in relation to performance in the reference condition. The primary CLCG of 0.078 represents an approximately 17\% reduction relative to the English Token-F1 score of 0.456. This is a relative reduction in the observed evaluation score, not a claim that the models understood 17\% less content or answered 17\% fewer questions correctly. The experimental design held source content, question, reference answer, model, prompt, context, and metric constant across parallel language conditions, while the executed estimator contrasted aggregate scored-row means. The reduction therefore represents a systematic controlled difference rather than an unadjusted comparison between unrelated benchmark conditions.

The result was also not driven by a single system. All five models exhibited a positive net CLCG, and their respective confidence intervals excluded zero. The magnitude varied across models, but none eliminated the difference among English, Portuguese, and the target languages. This suggests that the phenomenon is not specific to a particular architecture, provider, or model family.

The robustness analyses make an explanation based solely on a particular analytical choice unlikely. The direction of the effect persisted across automatic metrics, aggregation procedures, models, depth levels, and study subsets; an independent depth re-labeling reproduced the ordered slope, the non-monotonic profile, and its \texttt{L2} peak (Supplementary Material Section SM11); and the gap did not shrink when the context was narrowed from the full article to the linked paragraphs---if anything it grew, indicating the gap is not an artifact of locating evidence in a long document. Language ordering also showed high stability across repetitions and a strong association between the results obtained in the primary domain and the FLORES corpus.

Together, this evidence supports interpreting the CLCG as a systematic property of the behavior of the evaluated models: competence demonstrated in English does not transfer fully when the same content is presented in languages with lower resource availability.

This conclusion is more limited than claiming that the models ``do not understand'' these languages. The study measures performance on question-answering comprehension tasks rather than providing direct access to internal cognitive states. A positive CLCG indicates that the operationally observable ability to extract, preserve, and express relevant information varies across comparable language conditions.

The \texttt{closed\_book} analysis helps qualify this interpretation. The models derived greater benefit from the passage in English than in several lower-resource conditions, although the Joshi class means were not monotonic. The gap therefore does not appear to arise solely from differences in the prior knowledge required to answer. It is also associated with the amount of useful information the model can retrieve from the provided context.

The error analysis points in the same direction. The most frequent form of deterioration was not a response entirely disconnected from the passage, but the transformation of a correct response into a partial or incomplete one. This pattern suggests a gradual loss of semantic coverage: the model preserves some of the content but fails to retrieve one or more elements required to answer completely.

The combination of lower context gain, more frequent omissions, and an association with resource availability provides a coherent explanation for the observed phenomenon. In languages with more limited representation, the model appears to form a less complete or less usable representation of the presented content. This interpretation remains mechanistic and associative rather than causal because the study does not directly observe the systems' training data or internal states.

The primary result, therefore, is not merely that average performance decreases in certain languages. The language of presentation changes the competence that would be attributed to the same model if it were evaluated exclusively in English. Monolingual evaluations may consequently overestimate the comprehension ability available to users who interact with the models in lower-resource languages.

\begin{figure}
\centering
\pandocbounded{\includegraphics[width=\linewidth,keepaspectratio]{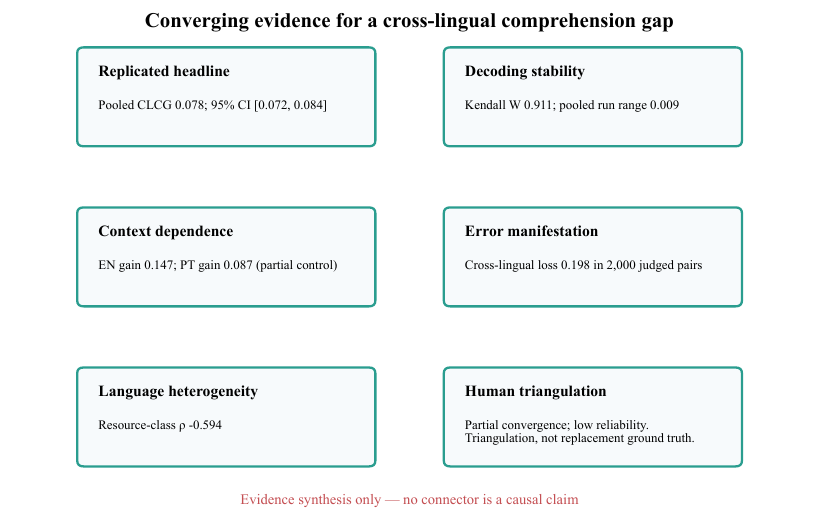}}
\caption{Synthesis of the executed CLCG evidence chain. Six blocks summarize the replicated headline, decoding stability, partial context diagnostic, automatic error pattern, language-level resource association, and limited human triangulation. The arrangement denotes converging evidence only and makes no causal claim.}
\label{fig:evidence-synthesis}
\end{figure}

Figure~\ref{fig:evidence-synthesis} presents a non-causal conceptual synthesis of the evidence supporting this interpretation.

\subsubsection{Human Triangulation: Partial Convergence and Limits to Validity}\label{human-triangulation-partial-convergence-and-limits-to-validity}

Human evaluation did not reproduce the CLCG as a second, fully reliable scale, but it provided convergent evidence at three levels. First, adequacy and factual coverage were positively correlated with the automatic score. Second, Joshi classes with larger CLCG values tended to receive lower human adequacy ratings, although this class-level association was imprecise. Third, in the human paired arm, which directly compared two responses for the same item and model, the response from the higher-resource condition was preferred in 61.6\% of decisive judgments. This human pairwise task is stricter than the aggregate micro-mean contrast used by the executed H1 estimator.

These signals make it less plausible that the CLCG is merely an idiosyncratic artifact of Token-F1. At the same time, the correlations were modest and inter-rater agreement was low. The human evidence therefore does not warrant treating the automatic metrics as calibrated substitutes for human judgment or interpreting small differences between responses as psychometrically stable facts.

The low agreement appears to reflect a combination of task subjectivity, responses that were often partial but plausible, and the difficulty of distinguishing adequacy, equivalence, attribution, and factual coverage in rapid crowdsourced judgments. It may also explain why the paired evidence was clearer than the absolute scales. Selecting the better of two comparable responses requires less individual calibration than assigning consistent absolute scores.

The sensitivity analyses support retaining these judgments in the primary analysis rather than retrospectively redefining the sample. Excluding all affected items produced a selectively simpler subset with fewer atomic facts, shorter gold answers, lower adequacy ratings, and higher factual recall. Treating that subset as the primary sample would therefore introduce selection rather than simply remove noise.

The disagreement between humans and the two automatic error judges further supports restricting the taxonomy to a descriptive role. The automatic judges identify useful aggregate patterns, such as omissions and unsupported additions, but their fine-grained categories do not constitute human-validated labels.

The evidence supporting the study's validity is therefore asymmetric. The existence and general direction of the gap are supported by multiple automatic analyses, cross-domain replication, stability, context gain, and paired human preference. The precise interpretation of its magnitude and failure categories, however, remains dependent on the measurement instrument. The CLCG should be understood as a robust operational estimator of relative performance loss, not as a comprehensive psychometric measure of human comprehension.

\subsection{Data and Code Availability}\label{data-and-code-availability}

The materials underlying this study are publicly available through two versioned Zenodo records. ParallelQA-18: Multilingual Parallel QA Predictions and Human Validation Dataset, version 3.5.0 (\url{https://doi.org/10.5281/zenodo.21815370}), contains the releasable collected bases (da Silva \& Eicher, 2026b). These include reconstruction manifests, identifiers, generated questions and gold answers, a hash-based index of source-authored human questions, frozen model responses, automatic and model-judge evaluations, and deidentified human-validation data. Copyrighted article bodies and source-authored question text are not redistributed in this record.

The software needed to reconstruct the copyrighted source materials locally is available separately as ParallelQA-18 Builder: Reconstruction and Evaluation Toolkit, version 3.6.0 (\url{https://doi.org/10.5281/zenodo.21816774}; da Silva \& Eicher, 2026a). The same Builder deposit archives the SPARQL retrieval code and frozen Wikidata language-metadata extract underlying Table~\ref{tab:language-inventory} (da Silva \& Eicher, 2026c). Paper-facing analysis code that consumes the Dataset and Builder outputs to regenerate \texttt{study\_results.json}, the analytical source of truth, is publicly available at \url{https://github.com/rafa-rodriguess/CLCG_pub}. The two Zenodo releases have separate persistent concept DOIs, allowing the dataset and reconstruction software to be cited independently.

\nocite{*}
\bibliographystyle{compling}
\bibliography{paper_clcg}

\end{document}